\documentclass[letterpaper]{article} 
\usepackage{aaai2026}  
\usepackage{times}  
\usepackage{helvet}  
\usepackage{courier}  
\usepackage[hyphens]{url}  
\usepackage{graphicx} 
\usepackage{natbib}  
\usepackage{caption} 
\usepackage{algorithm}
\usepackage{algorithmic}
\usepackage{xspace}

\usepackage{url}            
\usepackage{booktabs}       
\usepackage{amsfonts}       
\usepackage{nicefrac}       
\usepackage{microtype}      
\usepackage{xcolor}         
\usepackage{amsmath}

\newcommand{\ours}{\textsc{MedAlign}\xspace}
\usepackage{makecell}
\usepackage{tabularx}
\usepackage{multirow}
\usepackage[table]{xcolor}
\definecolor{myhighlight}{RGB}{230, 230, 230}

\usepackage{newfloat}
\usepackage{listings}
\DeclareCaptionStyle{ruled}{labelfont=normalfont,labelsep=colon,strut=off} 
\floatstyle{ruled}
\newfloat{listing}{tb}{lst}{}
\floatname{listing}{Listing}
\title{Enhancing Medical Large Vision-Language Models via Alignment Distillation}

\author{
Aofei Chang$^1$,
Ting Wang$^2$,
Fenglong Ma$^1$\thanks{Corresponding author.}}

\affiliations{
$^1$Pennsylvania State University \\
$^2$Stony Brook University \\
\{aofei,fenglong\}@psu.edu, twang@cs.stonybrook.edu

}

\usepackage{bibentry}

\begin{document}

\maketitle

\begin{abstract}
Medical Large Vision-Language Models (Med-LVLMs) have shown promising results in clinical applications, but often suffer from hallucinated outputs due to misaligned visual understanding. In this work, we identify two fundamental limitations contributing to this issue: insufficient visual representation learning and poor visual attention alignment. To address these problems, we propose \textbf{\ours}, a simple, lightweight alignment distillation framework that transfers visual alignment knowledge from a domain-specific Contrastive Language-Image Pre-training (CLIP) model to Med-LVLMs. \ours introduces two distillation losses: a spatial-aware visual alignment loss based on visual token-level similarity structures, and an attention-aware distillation loss that guides attention toward diagnostically relevant regions. Extensive experiments on medical report generation and medical visual question answering (VQA) benchmarks show that \ours consistently improves both performance and interpretability, yielding more visually grounded outputs. 
\end{abstract}

\begin{links}
    \link{Code}{https://github.com/Aofei-Chang/MedAlign}
\end{links}

\section{Introduction}\label{sec:introduction}

Medical Large Vision-Language Models (Med-LVLMs), such as LLaVA-Med-1.5 and HuatuoGPT-Vision, have shown strong potential in clinical applications~\cite{li2024llava, chen2024huatuogpt, chexagent-2024, thawkar2023xraygpt, moor2023med}. However, recent studies~\cite{xia2024cares, gu2024medvh, chen2024detecting, chang2025medheval, wang2024recent} have revealed that these models often produce inaccurate or hallucinated responses that fail to faithfully reflect the input medical images. To the best of our knowledge, \textbf{no} existing work has proposed targeted methods to mitigate hallucinations specifically in Med-LVLMs. Existing hallucination mitigation strategies primarily focus on general-purpose LVLMs and follow two main directions: (1) enhancing visual grounding and reducing over-reliance on textual input through contrastive decoding -- applied at either the attention or input level~\cite{leng2024mitigating, favero2024multi, liu2025paying, tu2025attention, chen2025attention}; and (2) correcting attention biases, such as the overemphasis on background elements or ``register'' tokens among visual inputs~\cite{darcet2024vision, woo2024don, gong2024damro}. 
While these techniques may alleviate hallucinations, they do not explicitly improve the distribution of visual attention or ensure that the model focuses on clinically relevant regions. More critically, they overlook a key contributor to hallucination in Med-LVLMs: \textit{the quality of the learned visual representations.} 

\noindent\textbf{Preliminary Analysis.}
To address these limitations, we conduct a preliminary analysis to investigate two fundamental factors driving hallucinations in Med-LVLMs: (1) the quality of the visual representations learned by the model, and (2) the alignment of visual attention during generation.

\begin{figure*}[t]
  \centering
  \includegraphics[width=0.85\linewidth]{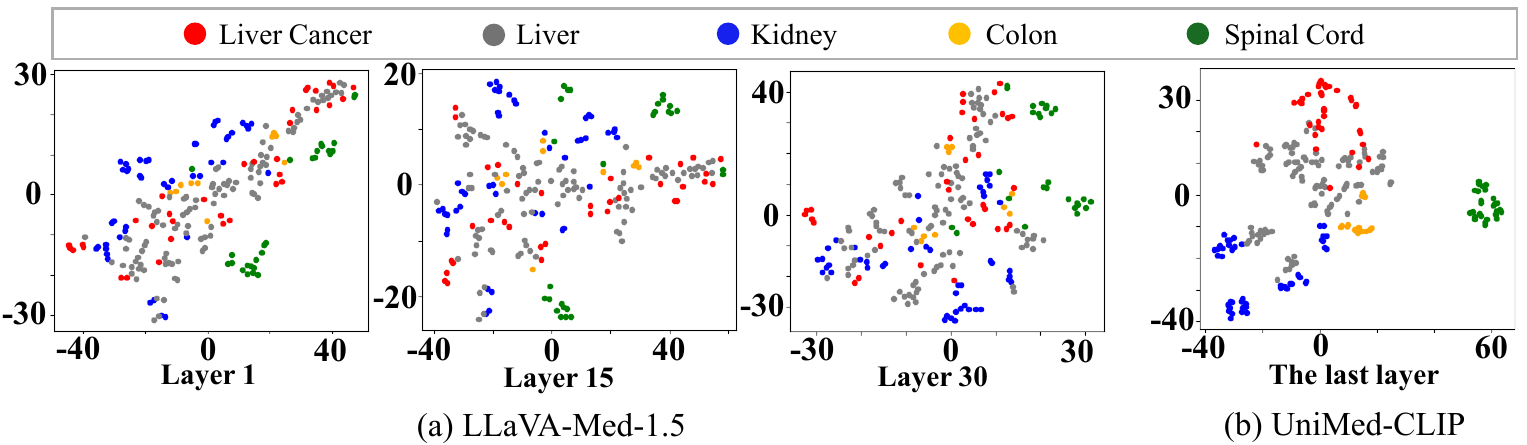}
  \caption{A $t$-SNE visualization of visual features derived from sampled abdominal CT scans using LLaVA-Med-1.5 and UniMed-CLIP.}
  \label{fig:preliminary_vis_feature}
\end{figure*}

\begin{figure*}[t]
  \centering
  \includegraphics[width=0.85\linewidth]{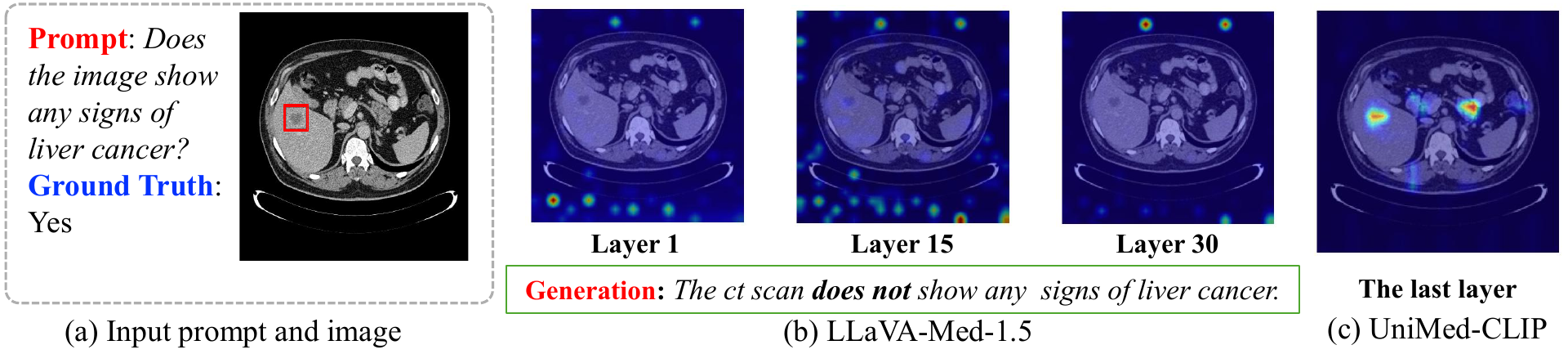}
  \caption{Attention visualizations from LLaVA-Med-1.5 and UniMed-CLIP on an abdominal CT scan. The red box (not part of the input) highlights the liver cancer region.}
  \label{fig:preliminary_vis_attention}
\end{figure*}

(1) Insufficient Visual Representation Learning. Unlike images in the general domain that contain diverse objects, medical images often feature recurring anatomical structures, such as the lungs, heart, and ribs in chest X-rays. Ideally, a well-trained Med-LVLM should learn similar representations for the same organ across different images. To evaluate the quality of visual representation learning in existing Med-LVLMs, we adopt LLaVA-Med-1.5 as a representative model. We randomly sample 100 abdominal CT scans from the SLAKE~\cite{liu2021slake} dataset, which includes Region-of-Interest (RoI) annotations for image patches. For analysis, we extract the visual token representations from various layers of the Transformer-based large language model (LLM) used in LLaVA-Med-1.5 and visualize five key entities using $t$-SNE. The results, shown in Figure~\ref{fig:preliminary_vis_feature} (a), illustrate how visual representations evolve across layers. We can observe that LLaVA-Med-1.5 fails to clearly distinguish key entities in medical images, resulting in \textbf{entangled} and \textbf{dispersed} visual representations. For example, the representations of ``\textit{liver cancer}'' are heavily mixed with those of ``\textit{liver}'' and other nearby organs. These results suggest that current Med-LVLMs exhibit insufficient visual representation learning, particularly for clinically critical concepts, which may potentially leads to poor visual reasoning and increased risk of hallucinations.

(2) Visual Attention Misalignment. A well-trained Med-LVLM should understand both the input image and the corresponding text prompt and assign higher attention weights to image regions relevant to the medical concepts mentioned in the prompt. However, as previously discussed, the issue of insufficient visual representation learning in Med-LVLMs leads to a secondary problem in the LLM component: visual attention misalignment. As illustrated in Figure~\ref{fig:preliminary_vis_attention} (b), we analyze a specific prompt, ``\textit{Does the image show any signs of liver cancer}'', by visualizing the average visual attention weights across $H$ attention heads in the $l$-th layer. The visualization reveals that the high attention weights are not aligned with the relevant region (highlighted by the red box in the input image), leading the model to generate a hallucinated response, ``\textit{The ct scan  \textbf{does not} show any signs of liver cancer.}'', despite the ground truth being ``\textit{Yes}''.

\noindent\textbf{Motivations of Incorporating Domain-specific Guidance.}
We hypothesize that the general CLIP-based image encoder used in Med-LVLMs is a primary contributor to both insufficient visual representation learning and visual attention misalignment. Leveraging a more domain-adapted CLIP-based model, such as UniMed-CLIP~\cite{khattak2024unimed}, which is trained on large-scale medical image–text data, could significantly enhance Med-LVLM performance. 
To explore this, we use UniMed-CLIP (large) as an example, which takes an image–text pair $(I, P)$ as input. The output of the image encoder consists of a concatenation of the $[\text{cls}]$ representation $\mathbf{E}_{[\text{cls}]} \in \mathbb{R}^b$ and patch-wise visual representations $\mathbf{E}_v \in \mathbb{R}^{M \times b}$, where $M$ denotes the number of image patches and $b$ is the hidden dimension.

As shown in Figure~\ref{fig:preliminary_vis_feature} (b), the visual features $\mathbf{E}_v$ extracted from the final layer of UniMed-CLIP exhibit improved semantic separation across medical concepts, suggesting more structured and discriminative visual representations. In addition, we extract the attention map $\mathbf{E}_a \in \mathbb{R}^M$, representing attention weights between the $[\text{cls}]$ token and the $M$ image patches from the last layer of the visual encoder. The visualization in Figure~\ref{fig:preliminary_vis_attention} (c) shows that UniMed-CLIP achieves stronger visual grounding by attending more precisely to diagnostically relevant regions. Together, these observations indicate that UniMed-CLIP provides more better representations and attention distributions than LLaVA-Med-1.5.

Given these findings, a straightforward strategy to enhance Med-LVLMs is to replace the original CLIP encoder with a domain-specific expert encoder like UniMed-CLIP. However, this requires re-training the visual projection layer and adapting the entire Med-LVLM to the new feature space using large-scale data, which is \textit{computationally intensive}. This limitation motivates the design of a lightweight, non-invasive approach to transfer alignment knowledge from expert CLIP models without fully replacing the original visual encoder.

\begin{figure*}[t]
  \centering
  \includegraphics[width=0.9\linewidth]{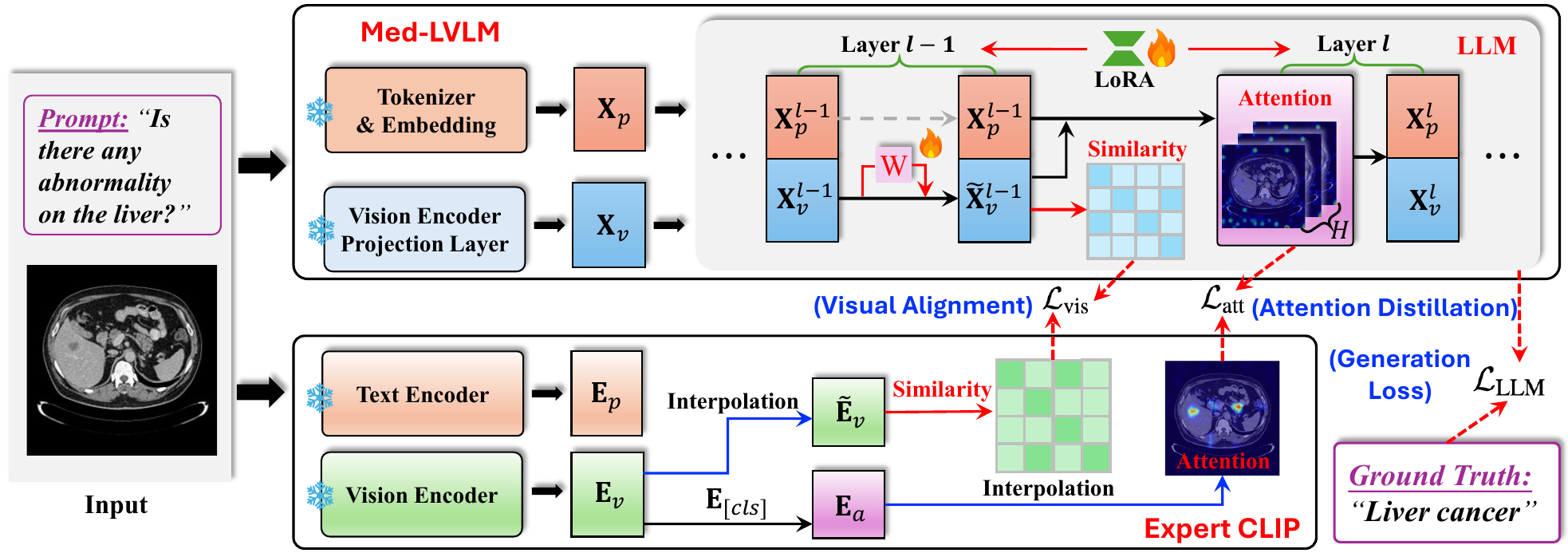}
  \caption{The overview of the proposed \ours that uses an expert CLIP model as a reference to guide the fine-tuning of Med-LVLMs via a spatial-aware visual alignment loss and an attention-aware distillation loss.}
  \label{fig:model_design}
\end{figure*}

\noindent\textbf{Our Approach.}
We propose \ours, a novel alignment distillation framework designed to enhance Med-LVLMs by transferring both visual representations and attention patterns from a domain-specific expert CLIP model. As illustrated in Figure~\ref{fig:model_design}, given an input image–prompt pair, we extract two alignment signals from the expert CLIP: (1) visual representations and (2) visual attention maps. These signals are then distilled into intermediate layers of the Med-LVLM to improve its alignment with clinically relevant content.
To enable lightweight and non-invasive integration, we introduce two core components. First, a \textbf{spatial-aware visual alignment loss} captures the pairwise similarity structure among image patches - reflected in the expert CLIP's visual features - and transfers it to the Med-LVLM’s internal representations. Second, an \textbf{attention-aware distillation loss} aligns the Med-LVLM’s attention distributions with those derived from the expert model. \ours does not require retraining or modifying visual encoders and provides a plug-and-play solution that integrates seamlessly into existing Med-LVLMs.

\noindent\textbf{Contributions.}
In summary, our work makes the following contributions:
(1) We conduct a detailed analysis of current Med-LVLMs and identify two key but underexplored sources of misalignment: misaligned visual representations and misaligned attention distributions.
(2) We introduce \ours, a novel alignment distillation framework that transfers alignment knowledge from expert CLIP models to Med-LVLMs without requiring model retraining or fine-grained supervision. The framework features two lightweight loss functions: spatial-aware visual alignment and attention-aware distillation, for effective and interpretable knowledge transfer.
(3) We validate \ours through extensive experiments on two report generation benchmarks and five medical VQA datasets, demonstrating consistent improvements over strong baselines in both task performance and visual interpretability. Our approach leads to better visual grounding and more clinically faithful outputs.
\section{Related Work}



While prior efforts such as A$^3$Tune~\cite{chang2025focus} and CoMT~\cite{Jiang2024CoMTCR} reduce hallucinations in Med-LVLMs, most focus on surface-level attention adjustments and overlook the core issue of multimodal misalignment. Since Med-LVLMs inherit architectures from general LVLMs, they exhibit similar hallucination behaviors, making several inference-time techniques transferable to the medical domain—such as improving visual grounding~\cite{leng2024mitigating,favero2024multi,liu2025paying,yuan2024helpd,liang2024mitigating,chen2025attention,tu2025attention}, correcting visual attention biases~\cite{woo2024don,gong2024damro}, and refining decoding~\cite{huang2024opera,chuang2023dola}. However, resolving modality misalignment in Med-LVLMs remains largely unsolved.


\section{The Proposed \ours}
\label{sec:method}

\subsection{Overview}
Based on the preliminary observations presented in the introduction, we propose a novel, lightweight, and straightforward distillation-based framework, named \ours, aimed at improving visual-text alignment in Med-LVLMs. \ours achieves this by transferring fine-grained alignment knowledge from a domain-specialized expert CLIP model to the target Med-LVLM. Specifically, two forms of knowledge extracted from the last layer of the expert CLIP image encoder are used to guide the learning process: the visual representations $\mathbf{E}_v \in \mathbb{R}^{M \times b}$ and the visual attention vector $\mathbf{E}_a \in \mathbb{R}^M$.

An overview of \ours is illustrated in Figure~\ref{fig:model_design}. Med-LVLMs take paired inputs of an image and a text prompt, denoted as $(I, P)$, similar to general LVLMs~\cite{liu2024visual}. The medical image $I$ is typically divided into $N$ patches and encoded by a CLIP-based image encoder and a visual projection layer, producing visual embeddings $\mathbf{X}_v \in \mathbb{R}^{N \times d}$, where $d$ is the token dimension used by the large language model (LLM). Meanwhile, the text prompt $P$ is processed through a text embedding layer, yielding $\mathbf{X}_p \in \mathbb{R}^{T \times d}$, where $T$ is the number of text tokens. The model then concatenates $\mathbf{X}_v$ and $\mathbf{X}_p$ and feeds the combined sequence into the $L$-layer Transformer-based LLM to generate the output. To address the issue of dispersed visual representations, we introduce a \textbf{spatial-aware visual alignment distillation} loss that transfers the relative similarity structure among image patches from the expert CLIP model to the Med-LVLM, using $\mathbf{E}_v$. To further align the attention distribution, we leverage $\mathbf{E}_a$ from the expert model to guide the visual attention learning in the Med-LVLM through an \textbf{attention-aware distillation} loss.
Importantly, both forms of alignment supervision are applied only at a designated layer $l$ within the Med-LVLM. Specifically, visual representation distillation is applied to the input of layer $l$, intervening at the output visual representations of layer $l-1$. This design enables a lightweight and non-invasive distillation process, avoiding full encoder replacement or costly end-to-end retraining.

\subsection{Spatial-aware Visual Alignment via $\mathbf{E}_v$}\label{sec:visual_alignment}
A straightforward approach to aligning the visual representations from layer $l-1$ of the Med-LVLM (i.e., $\mathbf{X}_v^{l-1} \in \mathbb{R}^{N \times d}$) with those from the expert model (i.e., $\mathbf{E}_v \in \mathbb{R}^{M \times b}$) is to minimize the distance between them. However, this is impractical due to the mismatch in dimensionality between the two matrices. Moreover, even if the dimensions were aligned, directly forcing the representations to match could adversely affect the LLM generation, leading to degraded performance.

\paragraph{Representation Rotation.}
To address these challenges, we introduce a trainable rotation matrix $\mathbf{W} \in \mathbb{R}^{d \times d}$ to adaptively adjust the Med-LVLM visual representations. The transformed representation is computed as:
\begin{equation}\label{eq:rotation}
\tilde{\mathbf{X}}_v^{l-1} = (\mathbf{I} + \mathbf{W})\mathbf{X}_v^{l-1},
\end{equation}
where $\mathbf{I} \in \mathbb{R}^{d \times d}$ is the identity matrix. This formulation allows for a lightweight and smooth transformation, gently steering $\mathbf{X}_v^{l-1}$ without drastic changes to the original representation space. The goal of this rotation is to bring visually similar concepts closer in the feature space, thus mitigating the representation dispersion issue. As demonstrated in Figure~\ref{fig:preliminary_vis_feature} (a), the visual representations suffer from poor semantic structure. Inspired by techniques such as word embedding steering in LLMs~\cite{han2023word}, which apply targeted modifications through small perturbations, our learnable matrix $\mathbf{W}$ serves as a controlled adjustment mechanism to enhance visual coherence in a minimally invasive manner.

\paragraph{Spatial-aware Visual Alignment.}
After obtaining the rotated visual representations $\tilde{\mathbf{X}}_v^{l-1} \in \mathbb{R}^{N \times d}$, we still face a dimensional mismatch with the expert visual features $\mathbf{E}_v \in \mathbb{R}^{M \times b}$. To resolve this, we first apply an interpolation function to adjust $\mathbf{E}_v$ such that it matches the number of patches of $\tilde{\mathbf{X}}_v^{l-1}$:
\begin{equation}
\tilde{\mathbf{E}}_v = \text{interpolate}(\mathbf{E}_v, (N, b)).
\end{equation}
However, directly minimizing the distance between $\tilde{\mathbf{X}}_v^{l-1}$ and $\tilde{\mathbf{E}}_v \in \mathbb{R}^{N \times b}$ may still negatively affect LLM performance due to rigid alignment. Instead, we propose leveraging the \textbf{pairwise similarity structure} among patch representations as a softer, more semantically meaningful alignment signal. Intuitively, patches depicting the same organ should have similar visual embeddings, resulting in higher similarity scores, while unrelated patches should be dissimilar.

To capture these structural relationships, we compute pairwise cosine similarity matrices for both the expert features $\tilde{\mathbf{E}}_v$ and the rotated Med-LVLM features $\tilde{\mathbf{X}}_v^{l-1}$:
\begin{equation}
\mathbf{S}^e_{i,j} = \cos(\tilde{\mathbf{E}}_v[i], \tilde{\mathbf{E}}_v[j]),   \text{and } \mathbf{S}^x_{i,j} = \cos(\tilde{\mathbf{X}}_v^{l-1}[i], \tilde{\mathbf{X}}_v^{l-1}[j]).
\end{equation}
The spatial-aware visual alignment distillation loss is then defined as the mean squared error between the two similarity matrices:
\begin{equation}
\mathcal{L}_{\text{vis}} = \frac{1}{N^2} \sum_{i=1}^N \sum_{j=1}^N \left(\mathbf{S}^e_{i,j} - \mathbf{S}^x_{i,j}\right)^2.
\end{equation}
This loss encourages the model to capture the same structural similarity patterns among image patches as the expert model, promoting semantically coherent and spatially aware visual representations without enforcing direct value matching.

\subsection{Attention-aware Alignment via $\mathbf{E}_a$}\label{sec:attention_alignment}

In addition to refining visual representations, Med-LVLMs also suffer from visual attention misalignment, where attention weights fail to properly focus on diagnostically relevant regions. In each Transformer layer $l$ of a Med-LVLM, $H$ attention heads are used, and each head $h$ produces a visual attention vector $\mathbf{M}^l_{a,h} \in \mathbb{R}^N$, where $N$ is the number of image patches. The average visual attention across all heads at layer $l$ is computed as
$\tilde{\mathbf{M}}_a^l = \frac{1}{H} \sum_{h=1}^H \mathbf{M}^l_{a,h}$.

However, aligning this attention distribution with the expert attention vector $\mathbf{E}_a \in \mathbb{R}^M$ is not straightforward due to a mismatch in patch numbers (i.e., $M$ may not be equal to $N$). To address this, we apply interpolation to resize $\mathbf{E}_a$ to match the Med-LVLM patch resolution, yielding an interpolated expert attention vector $\tilde{\mathbf{E}}_a \in \mathbb{R}^N$.

To guide attention learning, we treat $\tilde{\mathbf{E}}_a$ as the teacher and $\tilde{\mathbf{M}}_a^l$ as the student. We then define the attention-aware alignment loss using the Kullback–Leibler (KL) divergence:
\begin{equation}
\label{eq:kl}
\mathcal{L}_{\text{att}} = \text{KL}(\tilde{\mathbf{E}}_a || \tilde{\mathbf{M}}_a^l),
\end{equation}
where both distributions are normalized using softmax. This loss encourages the Med-LVLM to mimic the expert model's attention focus, promoting more accurate grounding of visual information and improving the quality of generated responses.

\subsection{Final Objective}
\label{sec:method_joint}

We implement \ours using parameter-efficient fine-tuning based on LoRA~\cite{hu2021lora}, applied to all linear layers in the LLM of the Med-LVLM. During fine-tuning, we jointly optimize the proposed alignment losses alongside the standard language modeling objective $\mathcal{L}_{\text{LLM}}$, defined as:
\begin{equation}
    \mathcal{L}_{\text{LLM}}  = -\sum_{t=1}^T \log p_t(y_t \mid \mathbf{X}_v, \mathbf{X}_p, y_{<t}; \Theta, \Phi),
\end{equation}
where $\Theta$ denotes the \textit{frozen} parameters of the base LLM, and $\Phi$ denotes the trainable parameters introduced by LoRA and our designed alignment modules. The final training target is:
\begin{equation}\label{eq:final}
    \mathcal{L} = \mathcal{L}_{\text{LLM}} + \alpha \mathcal{L}_{\text{vis}} + \beta\mathcal{L}_{\text{att}},
\end{equation}
where $\alpha$ and $\beta$ are hyperparameters that balance alignment distillation with the language modeling loss $\mathcal{L}_{\text{LLM}}$.

\section{Experiments}

\begin{table*}[t]
\centering
\footnotesize 
\begin{tabular}{p{1.2cm}@{\hskip 6pt}|@{\hskip 6pt}p{1.6cm}@{\hskip 8pt}|c@{\hskip 5pt}c@{\hskip 5pt}c@{\hskip 5pt}c@{\hskip 7pt}c@{\hskip 4pt}c@{\hskip 3pt}c@{\hskip 5pt}c@{\hskip 5pt}c@{\hskip 5pt}c@{\hskip 5pt}|c}
\toprule
\multirow{2}{*}{\textbf{Dataset}} & \multirow{2}{*}{\textbf{Metric}} & \multicolumn{11}{c}{\textbf{Method}}\\
\cline{3-13}
&  
& Greedy 
& Beam
& Nucleus
& VCD
& DoLa
& OPERA
& AVISC
& M3ID
& DAMRO
& PAI
& \cellcolor{myhighlight}{\ours} \\
\midrule
\multirow{7}{*}{\textbf{IU-Xray}} 
& {BLEU}        & 9.34 & 10.21 & 8.19 & 9.10 & 9.03 & 10.01 & 7.47 & 8.35 & 9.10 & 6.92 & \cellcolor{myhighlight}\textbf{10.73} \\
& {ROUGE-L}     & 28.17 & 28.64 & 26.28 & 27.80 & 27.50 & 28.57 & 24.78 & 27.05 & 27.80 & 24.36 & \cellcolor{myhighlight}\textbf{29.10} \\
& {METEOR}      & 31.76 & 34.23 & 30.72 & 32.13 & 31.24 & 34.10 & 30.88 & 32.17 & 32.13 & 30.84 & \cellcolor{myhighlight}\textbf{36.30} \\
& {BERTScore}   & 88.53 & 88.60 & 88.19 & 88.36 & 88.39 & 88.51 & 87.77 & 88.19 & 88.36 & 87.44 & \cellcolor{myhighlight}\textbf{88.67} \\
& {CheXbert}    & 55.16 & 55.84 & 53.86 & 54.34 & 53.89 & 55.01 & 52.11 & 53.78 & 54.34 & 50.09 & \cellcolor{myhighlight}\textbf{56.27} \\
& {RadGraph}    & 21.86 & 22.47 & 20.17 & 21.65 & 21.04 & 22.59 & 19.71 & 21.19 & 21.65 & 18.26 & \cellcolor{myhighlight}\textbf{23.51} \\
& {RaTEScore}   & 58.66 & 59.78 & 58.29 & 58.29 & 57.84 & 59.33 & 56.49 & 57.86 & 58.29 & 55.16 & \cellcolor{myhighlight}\textbf{60.49} \\
\midrule
\multirow{7}{*}{\makecell{\textbf{MIMIC-}\\\textbf{CXR}}} 
& {BLEU}        & 4.22 & 4.16 & 3.61 & 3.73 & 3.98 & 4.04 & 3.55 & 3.74 & 3.73 & 3.87 & \cellcolor{myhighlight}\textbf{4.76} \\
& {ROUGE-L}     & 18.11 & 18.26 & 16.93 & 17.25 & 17.37 & 18.03 & 16.15 & 17.08 & 17.25 & 16.92 & \cellcolor{myhighlight}\textbf{19.32} \\
& {METEOR}      & 20.54 & 19.79 & 19.56 & 20.20 & 20.90 & 19.80 & 19.71 & 20.74 & 20.20 & 20.86 & \cellcolor{myhighlight}\textbf{22.02} \\
& {BERTScore}   & 85.79 & 85.84 & 85.35 & 85.53 & 85.47 & 85.73 & 84.70 & 85.25 & 85.53 & 84.59 & \cellcolor{myhighlight}\textbf{86.02} \\
& {CheXbert}    & 27.49 & 27.56 & 26.72 & 25.80 & 26.33 & 28.15 & 25.84 & 26.91 & 25.80 & 26.80 & \cellcolor{myhighlight}\textbf{29.53} \\
& {RadGraph}    & 11.75 & 10.85 & 9.96 & 10.69 & 11.39 & 10.60 & 10.09 & 10.67 & 10.69 & 10.57 & \cellcolor{myhighlight}\textbf{12.82} \\
& {RaTEScore}   & 43.39 & 42.38 & 41.66 & 42.54 & 42.71 & 41.93 & 42.07 & 42.37 & 42.54 & 42.48 & \cellcolor{myhighlight}\textbf{44.96} \\
\bottomrule
\end{tabular}
\caption{Report generation results of HuatuoGPT-Vision-7B fine-tuned with LoRA. Best results are highlighted in \textbf{bold}.}
\label{tab:main_exp_report_01}
\end{table*}

\begin{table*}[t]
\centering

\footnotesize
\begin{tabular}{p{1.2cm}@{\hskip 6pt}|@{\hskip 6pt}p{1.6cm}@{\hskip 8pt}|c@{\hskip 5pt}c@{\hskip 5pt}c@{\hskip 5pt}c@{\hskip 7pt}c@{\hskip 4pt}c@{\hskip 3pt}c@{\hskip 5pt}c@{\hskip 5pt}c@{\hskip 5pt}c@{\hskip 5pt}|c}

\toprule
\multirow{2}{*}{\textbf{Dataset}} & \multirow{2}{*}{\textbf{Metric}} & \multicolumn{11}{c}{\textbf{Method}}\\
\cline{3-13}
&  
& Greedy 
& Beam
& Nucleus
& VCD
& DoLa
& OPERA
& AVISC
& M3ID
& DAMRO
& PAI
& \cellcolor{myhighlight}{\ours} \\
\midrule
\multirow{7}{*}{\textbf{IU-Xray}} 
& {BLEU}        & 9.36 & 9.54 & 7.80 & 8.83 & 8.93 & 9.23 & 5.57 & 8.44 & 8.21 & 8.52 & \cellcolor{myhighlight}\textbf{10.31} \\
& {ROUGE-L}     & 27.57 & 28.41 & 26.72 & 27.36 & 26.94 & 27.48 & 21.71 & 26.21 & 25.77 & 26.97 & \cellcolor{myhighlight}\textbf{29.01} \\
& {METEOR}      & 27.91 & 35.40 & 30.33 & 31.77 & 25.74 & 34.17 & 26.84 & 30.86 & 30.58 & 28.63 & \cellcolor{myhighlight}\textbf{35.22} \\
& {BERTScore}   & 88.55 & 88.45 & 88.28 & 88.30 & 88.42 & 88.17 & 87.34 & 88.20 & 88.09 & 88.42 & \cellcolor{myhighlight}\textbf{88.66} \\
& {CheXbert}    & 52.44 & 53.70 & 52.73 & 51.86 & 52.27 & 51.65 & 47.32 & 51.13 & 50.10 & 52.22 & \cellcolor{myhighlight}\textbf{55.62} \\
& {RadGraph}    & 21.28 & 22.43 & 20.85 & 22.02 & 20.63 & 21.37 & 16.87 & 20.77 & 22.33 & 20.99 & \cellcolor{myhighlight}\textbf{23.29} \\
& {RaTEScore}   & 58.61 & 59.65 & 57.84 & 58.93 & 58.10 & 57.89 & 53.66 & 59.37 & 57.31 & 58.21 & \cellcolor{myhighlight}\textbf{59.99} \\
\midrule
\multirow{7}{*}{\makecell{\textbf{MIMIC-}\\\textbf{CXR}}} 
& {BLEU}        & 3.50 & 3.66 & 3.48 & 3.74 & 3.48 & 3.56 & 3.31 & 3.14 & 3.42 & 3.63 & \cellcolor{myhighlight}\textbf{4.51} \\
& {ROUGE-L}     & 16.49 & 16.85 & 16.35 & 16.88 & 16.45 & 16.77 & 16.36 & 16.13 & 16.63 & 16.65 & \cellcolor{myhighlight}\textbf{18.43} \\
& {METEOR}      & 18.71 & 20.68 & 18.93 & 19.03 & 18.66 & 20.10 & 18.64 & 18.52 & 18.87 & 18.61 & \cellcolor{myhighlight}\textbf{20.80} \\
& {BERTScore}   & 85.54 & 85.51 & 85.50 & 85.56 & 85.54 & 85.46 & 85.48 & 85.39 & 85.48 & 85.60 & \cellcolor{myhighlight}\textbf{86.00} \\
& {CheXbert}    & 23.43 & 25.00 & 22.21 & 22.98 & 23.34 & 24.31 & 23.31 & 22.42 & 23.30 & 24.51 & \cellcolor{myhighlight}\textbf{25.67} \\
& {RadGraph}    & 9.63 & 9.91 & 9.27 & 9.56 & 9.52 & 9.81 & 9.02 & 9.15 & 9.46 & 9.60 & \cellcolor{myhighlight}\textbf{10.92} \\
& {RaTEScore}   & 40.49 & 41.46 & 40.08 & 40.93 & 40.49 & 41.33 & 40.36 & 39.74 & 40.80 & 40.49 & \cellcolor{myhighlight}\textbf{42.03} \\
\bottomrule
\end{tabular}
\caption{Performance on report generation benchmarks using LLaVA-Med-1.5 fine-tuned with LoRA.}
\label{tab:main_exp_report_02}
\end{table*}

\subsection{Experimental Setups}

\textbf{Med-LVLMs and Expert CLIPs.}
We evaluate our method on two representative Med-LVLMs: LLaVA-Med-1.5 and HuatuoGPT-Vision-7B, with Beam search as the default decoding strategy. In our main experiments, we use UniMed-CLIP (ViT-L/14, 336px) as the expert CLIP model. To further examine the impact of expert model selection, we also investigate additional configurations 
including UniMed-CLIP (ViT-B/16, 224px) and BiomedCLIP~\cite{zhang2023biomedclip}.

\textbf{Evaluation Tasks and Datasets.}
We evaluate our method on two core tasks in medical applications of Med-LVLMs: medical report generation and medical visual question answering (VQA). For the report generation task, we use MIMIC-CXR~\cite{johnson2019mimic} and IU-Xray~\cite{demner2016preparing}. For the VQA task, we adopt a diverse set of benchmark datasets, including SLAKE~\cite{liu2021slake}, VQA-RAD~\cite{lau2018dataset}, PathVQA~\cite{he2020pathvqa}, IU-Xray, and OmniMedVQA~\cite{hu2024omnimedvqa}.  


\textbf{Baselines.}
We compare against popular hallucination mitigation methods, including decoding strategies and contrastive decoding techniques.
The {decoding} baselines include Greedy decoding, Nucleus sampling, and Beam search.
For {contrastive decoding} methods, we evaluate against recent techniques including VCD~\cite{leng2024mitigating}, OPERA~\cite{huang2024opera}, DoLa~\cite{chuang2023dola}, 
AVISC~\cite{woo2024don}, M3ID~\cite{favero2024multi}, DAMRO~\cite{gong2024damro} and
PAI~\cite{liu2025paying}.

\begin{table*}[t]
\centering
\footnotesize
\begin{tabular}{c|l|cc|cc|cc|c|c}
\toprule
\multirow{2}{*}{\textbf{Med-LVLM}} & \multirow{2}{*}{\textbf{Method}} & \multicolumn{2}{c|}{\textbf{SLAKE}} & \multicolumn{2}{c|}{\textbf{VQA-RAD}} & \multicolumn{2}{c|}{\textbf{PathVQA}} & \multicolumn{1}{c|}{\textbf{IU-Xray}} & \multicolumn{1}{c}{\textbf{OmniMedVQA}} \\ \cline{3-10}
 &  & \textbf{Open} & \textbf{Close} & \textbf{Open} & \textbf{Close} & \textbf{Open} & \textbf{Close} & \textbf{Close} & \textbf{Close} \\
\midrule

\multirow{11}{*}{\textbf{\rotatebox{90}{\makecell{HuatuoGPT-Vision\\ + LoRA}}}}

& Greedy & 85.57 & 90.14 & 41.37 & 76.77 & 37.08 & 93.31 & 85.33 & 91.33  \\
& Beam & 86.03 & 90.14 & 43.75 & 76.77 & 37.16 & 93.16 & 85.33 & 91.33  \\
& Nucleus & 84.75 & 90.42 & 38.98 & 77.56 & 34.03 & 93.45 & 85.59 & 90.50  \\
& VCD & 84.73 & 89.58 & 40.81 & 77.95 & 34.97 & 93.01 & 85.08 & 90.50 \\
& DoLa & 85.30 & 89.86 & 42.08 & 77.95 & 35.92 & 91.71 & 85.71 & 91.33 \\
& OPERA & 86.01 & 90.14 & 43.57 & 76.77 & 37.40 & 93.16 & 85.20 & 91.37 \\
& AVISC & 84.16 & 91.27 & 38.97 & 78.35 & 35.48 & 93.34 & 85.33 & 90.76 \\
& M3ID & 85.00 & 89.86 & 41.79 & 77.95 & 35.93 & 93.39 & 85.33 & 91.52 \\
& DAMRO & 84.73 & 89.58 & 40.81 & 77.95 & 34.97 & 93.01 & 85.08 & 90.50  \\
& PAI & 84.40 & 90.14 & 41.83 & 77.17 & 35.87 & 92.95 &  85.71 & 90.99 \\




& \cellcolor{myhighlight}{\ours} & \cellcolor{myhighlight}\textbf{86.85} & \cellcolor{myhighlight}\textbf{92.39} & \cellcolor{myhighlight}\textbf{43.75} & \cellcolor{myhighlight}\textbf{78.74} & \cellcolor{myhighlight}\textbf{38.49} & \cellcolor{myhighlight}\textbf{93.63} & \cellcolor{myhighlight}\textbf{86.22} & \cellcolor{myhighlight}\textbf{93.60}\\

\midrule

\multirow{11}{*}{\textbf{\rotatebox{90}{\makecell{LLaVA-Med-1.5 \\ + LoRA}}}} 

& Greedy  & 82.97  & 88.45 & 36.70 & 74.41 & 37.98 & 93.22 & 84.69 & 91.03  \\
& Beam & 83.27 & 88.45 & 36.95 & 74.41 & 38.32 & 92.98 & 84.82 & 90.99\\
& Nucleus & 82.75 & 86.76 & 36.58 & 73.62 & 35.13 & 92.95 & 85.20& 90.69 \\
& VCD & 82.89 & 86.76 & 35.12 & 73.62 & 35.76 & 92.92 & 85.59 &90.69\\
& DoLa & 82.97 & 88.45 & 36.70 & 74.41 & 37.95 & 93.22 & 84.69 &91.03 \\
& OPERA & 83.07 & 88.45 & 36.91 & 74.80 & 38.64 & 93.28 & 84.69 &91.03 \\
& AVISC & 83.26 & 87.32 & 36.58 & 74.41 & 36.52 & 93.34 & 84.82 & 91.14 \\
& M3ID & 83.20 & 87.04 & 35.90 & 74.02 & 35.53 & 92.63 & 84.44 & 90.69\\
& DAMRO & 82.89 & 86.76 & 35.12 & 73.62  & 35.76 & 92.92 & 85.59 & 90.69  \\
& PAI & 83.50 & \textbf{89.01} & 34.54 & 74.41 & 37.46 & 93.24 & 84.69 & 90.92\\



& \cellcolor{myhighlight}{\ours} & \cellcolor{myhighlight}\textbf{84.85} & \cellcolor{myhighlight}\textbf{89.01} & \cellcolor{myhighlight}\textbf{39.62} & \cellcolor{myhighlight}\textbf{74.80} & \cellcolor{myhighlight}\textbf{38.65} & \cellcolor{myhighlight}\textbf{93.51} & \cellcolor{myhighlight}\textbf{85.84} & \cellcolor{myhighlight}\textbf{93.38}\\

\bottomrule
\end{tabular}
\caption{
Performance comparison on medical VQA benchmarks using HuatuoGPT-Vision-7B and LLaVA-Med-1.5. ``Open'' denotes open-ended answers; ``Close'' refers to close-ended (e.g., yes/no or multiple-choice) responses. 
}
\label{tab:main_exp_vqa}
\end{table*}

\textbf{Metrics.}
For the \textit{medical report generation} task, we adopt standard text generation metrics, including BLEU~\cite{papineni2002bleu}, ROUGE-L~\cite{lin2004rouge}, METEOR~\cite{banerjee2005meteor}, and BERTScore~\cite{zhang2019bertscore}. Additionally, we include domain-specific evaluation metrics tailored to medical report generation: CheXbert~\cite{smit2020combining}, RadGraph~\cite{jain2021radgraph}, and RaTEScore~\cite{zhao2024ratescore}. 
For the \textit{medical VQA} task, we follow the evaluation of LLaVA-Med~\cite{li2024llava}, reporting Accuracy for close-ended VQA and Recall for open-ended VQA.

\subsection{Medical Report Generation Results}
\label{sec:report_results}
Table~\ref{tab:main_exp_report_01} and Table~\ref{tab:main_exp_report_02} present the evaluation of \ours on medical report generation using HuatuoGPT-Vision-7B and LLaVA-Med-1.5, respectively. We report both traditional generation metrics (e.g., BLEU, ROUGE-L) and domain-specific metrics such as RaTEScore. All baselines are applied to LoRA-tuned models, as the original Med-LVLMs perform poorly on report generation tasks. It is shown that \ours outperforms all baselines across both models and most evaluation metrics. Notably, with HuatuoGPT-Vision-7B, \ours achieves the best scores on all metrics, including substantial improvements in report-specific metrics, demonstrating stronger generation quality.

\subsection{Medical VQA Results}
Table~\ref{tab:main_exp_vqa} reports results on five medical VQA benchmarks using HuatuoGPT-Vision-7B and LLaVA-Med-1.5. \ours consistently surpasses all baselines on both open- and close-ended questions, demonstrating strong and stable alignment across datasets. Although some methods excel in isolated cases (e.g., PAI and AVISC on SLAKE close-ended), their performance varies widely, revealing limited generalization. In contrast, \ours maintains balanced gains and shows particularly notable improvements on open-ended tasks—for example, on LLaVA-Med-1.5, +1.58 on SLAKE and +2.67 on VQA-RAD over the strongest baselines. This trend echoes its strong report-generation results, indicating that open-ended generation is especially sensitive to alignment quality and highlighting the effectiveness of our alignment distillation.

\begin{figure}
    \centering
    \includegraphics[width=0.8\columnwidth]{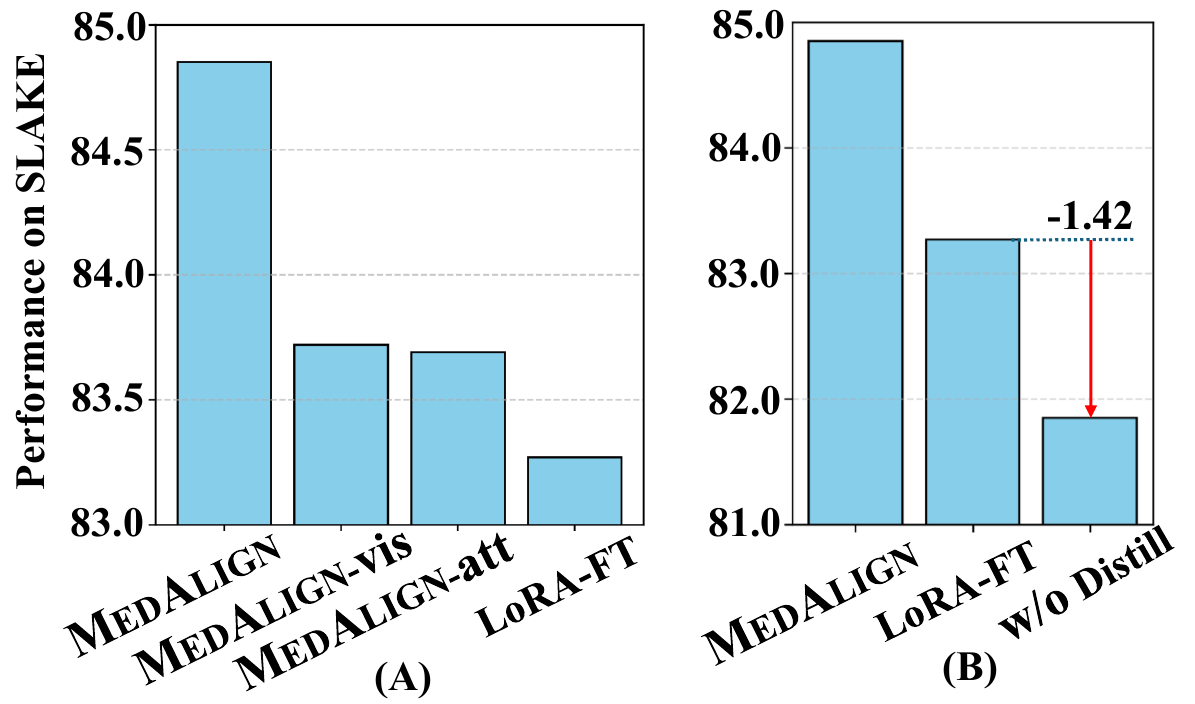}
    \caption{Ablation study on distillation design. (A) Impact of removing loss designs. (B) Performance when directly using UniMed-CLIP features without distillation (``w/o Distill'').}
    \label{fig:ablation_distill}
\end{figure}

\begin{figure*}[t]
  \centering
  \includegraphics[width=0.78\linewidth]{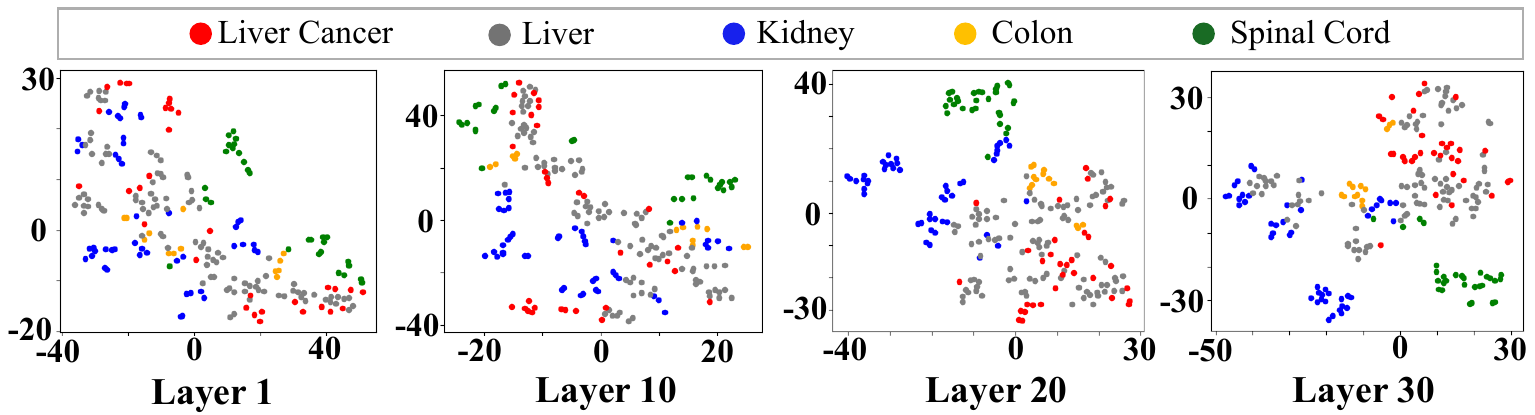}
  \caption{A $t$-SNE visualization of visual features from multiple layers of LLaVA-Med-1.5 after applying \ours.}
  \label{fig:case_study_vis_feature}
\end{figure*}
\begin{figure*}[t]
  \centering
  \includegraphics[width=0.85\linewidth]{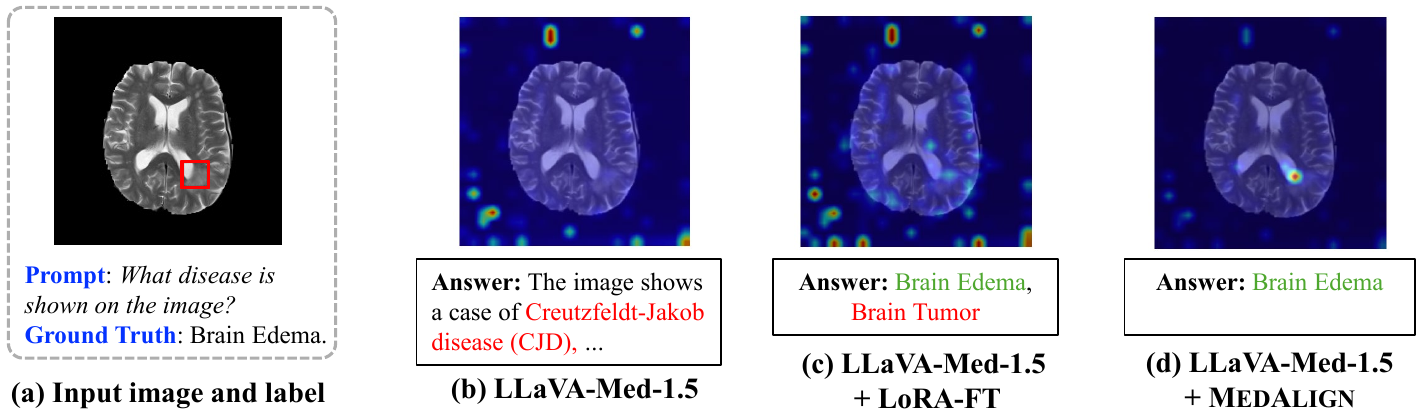}
  \caption{Comparison of overall attention distributions on the visual input averaged across all layers on a Brain MRI VQA example from SLAKE. Red text indicates hallucinated answers.}
  \label{fig:case_study_attn}
\end{figure*}

\subsection{Model Design Analysis}
Given that open-ended tasks better reflect the model's ability to interpret and describe medical images and are more sensitive to alignment quality, we focus our analysis primarily on the SLAKE open-ended subset.


\paragraph{Ablation Study on Distillation Design.}

Our proposed distillation framework employs two alignment loss terms, $\mathcal{L}_{\text{vis}}$ and $\mathcal{L}_{\text{att}}$, as defined in Eq.~\eqref{eq:final}. To evaluate the individual contributions of each component, we conduct an ablation study by selectively removing each loss term. Specifically, we denote the variants as \textsc{MedAlign}$-{\text{att}}$ (removing $\mathcal{L}_{\text{vis}}$) and \textsc{MedAlign}$-{\text{vis}}$ (removing $\mathcal{L}_{\text{att}}$). We also include a baseline using LoRA-based fine-tuning with Beam search, denoted as LoRA-FT.
As shown in Figure~\ref{fig:ablation_distill} (A), both loss terms individually improve performance over LoRA-FT, while combining them yields the best results, confirming their complementary benefits. In Figure~\ref{fig:ablation_distill} (B), directly using UniMed-CLIP features without distillation leads to performance degradation, likely due to feature mismatch. These results show the effectiveness of our distillation design in enhancing Med-LVLM performance without directly perturbing the feature space. 


\begin{figure}
    \centering
    \includegraphics[width=0.83\columnwidth]{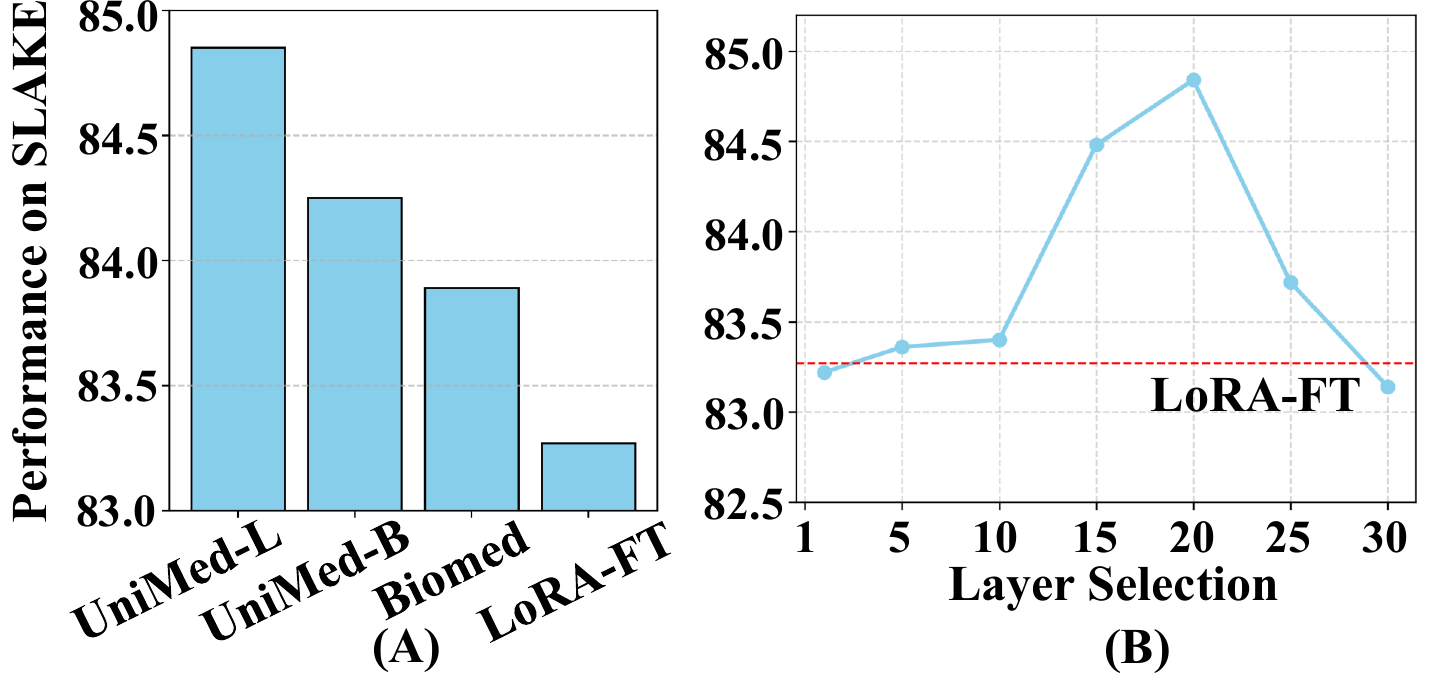}
  \caption{(A) Comparison of expert CLIP models. (B) Results of layer selection.}
    \label{fig:clip_layer_selection}
\end{figure}

\paragraph{Qualitative Analysis.}
To evaluate the effectiveness of our loss design, we examine changes in visual representations and attention after alignment distillation. For visual representations, we visualize $t$-SNE projections of features from layers 1–30 (Figure~\ref{fig:case_study_vis_feature}). Because alignment is applied at layer 20, feature separability improves noticeably from that layer onward, with enhancements even in earlier layers (e.g., layer 10), indicating backward propagation of alignment benefits. For visual attention, we visualize average attention on a Brain MRI VQA example (Figure~\ref{fig:case_study_attn}). Models trained with \ours produce more accurate answers and concentrate attention on clinically relevant regions, confirming the effectiveness of our attention alignment.

\subsection{Expert CLIP Model Selection}
\label{exp:clip_selection}
We use UniMed-CLIP (ViT-L/14, 336px), denoted \textbf{UniMed-L}, as our default expert. To assess expert choice and distillation generalizability, we also test UniMed-CLIP (ViT-B/16), \textbf{UniMed-B}, and BiomedCLIP (ViT-B/16), \textbf{Biomed}. Both UniMed-B and Biomed produce only 196 visual tokens—significantly fewer than the 576 used in LLaVA-Med-1.5 and HuatuoGPT-Vision-7B. To ensure compatibility, we apply interpolation on both the image representations and attention maps. 


As shown in Figure~\ref{fig:clip_layer_selection} (A), all expert CLIP models outperform LoRA-FT and prior baselines (Table~\ref{tab:main_exp_vqa}), with UniMed-L achieving the highest gains due to its finer token granularity. In contrast, UniMed-B and Biomed offer weaker supervision. These results demonstrate the flexibility of our framework and the value of strong visual alignment priors.

\subsection{Distillation Layer Selection}
In our experiments, we set the distillation layer to $l = 20$. To investigate the effect of layer selection, we varied the position of alignment. As shown in Figure~\ref{fig:clip_layer_selection} (B), applying \ours at early layers leads to performance degradation, likely due to disruption of low-level representations. Performance peaks in the middle layers, precisely where fine-grained multimodal interactions are known to occur~\cite{jiang2024devils, neo2024towards}. Applying the alignment losses at later layers also results in a drop in performance, likely due to limited representational flexibility in these stages.

\section{Conclusion}

In this work, we present \ours, a lightweight alignment distillation framework that enhances Med-LVLMs by transferring visual representation and attention alignment knowledge from expert medical CLIP models. Through designed distillation objectives, \ours improves visual grounding and output fidelity without requiring fine-grained annotations. Experiments on multiple benchmarks demonstrate consistent gains in both performance and interpretability, offering a practical path toward more reliable Med-LVLMs.
\section{Acknowledgments}
The authors thank the anonymous reviewers for their valuable comments and helpful suggestions. Dr. Ma is partially supported by the National Science Foundation under Grant No. 2238275 and the National Institutes of Health under Grant No. R01AG077016. Dr. Wang is partially supported by the National Science Foundation under Grant No. 2405136 and 2406572.

\bibliography{aaai2026}

@article{li2024llava,
  title={Llava-med: Training a large language-and-vision assistant for biomedicine in one day},
  author={Li, Chunyuan and Wong, Cliff and Zhang, Sheng and Usuyama, Naoto and Liu, Haotian and Yang, Jianwei and Naumann, Tristan and Poon, Hoifung and Gao, Jianfeng},
  journal={Advances in Neural Information Processing Systems},
  volume={36},
  year={2024}
}

@inproceedings{chen2024huatuogpt,
  title={Towards injecting medical visual knowledge into multimodal llms at scale},
  author={Chen, Junying and Gui, Chi and Ouyang, Ruyi and Gao, Anningzhe and Chen, Shunian and Chen, Guiming and Wang, Xidong and Cai, Zhenyang and Ji, Ke and Wan, Xiang and others},
  booktitle={Proceedings of the 2024 Conference on Empirical Methods in Natural Language Processing},
  pages={7346--7370},
  year={2024}
}

@inproceedings{thawkar2023xraygpt,
  title={XrayGPT: Chest radiographs summarization using large medical vision-language models},
  author={Thawakar, Omkar Chakradhar and Shaker, Abdelrahman M and Mullappilly, Sahal Shaji and Cholakkal, Hisham and Anwer, Rao Muhammad and Khan, Salman and Laaksonen, Jorma and Khan, Fahad},
  booktitle={Proceedings of the 23rd workshop on biomedical natural language processing},
  pages={440--448},
  year={2024}
}

@inproceedings{moor2023med,
  title={Med-flamingo: a multimodal medical few-shot learner},
  author={Moor, Michael and Huang, Qian and Wu, Shirley and Yasunaga, Michihiro and Dalmia, Yash and Leskovec, Jure and Zakka, Cyril and Reis, Eduardo Pontes and Rajpurkar, Pranav},
  booktitle={Machine Learning for Health (ML4H)},
  pages={353--367},
  year={2023},
  organization={PMLR}
}

@article{chang2025medheval,
  title={MedHEval: Benchmarking Hallucinations and Mitigation Strategies in Medical Large Vision-Language Models},
  author={Chang, Aofei and Huang, Le and Bhatia, Parminder and Kass-Hout, Taha and Ma, Fenglong and Xiao, Cao},
  journal={arXiv preprint arXiv:2503.02157},
  year={2025}
}

@inproceedings{huang2024opera,
  title={Opera: Alleviating hallucination in multi-modal large language models via over-trust penalty and retrospection-allocation},
  author={Huang, Qidong and Dong, Xiaoyi and Zhang, Pan and Wang, Bin and He, Conghui and Wang, Jiaqi and Lin, Dahua and Zhang, Weiming and Yu, Nenghai},
  booktitle={Proceedings of the IEEE/CVF Conference on Computer Vision and Pattern Recognition},
  pages={13418--13427},
  year={2024}
}

@inproceedings{
    chuang2023dola,
    title={DoLa: Decoding by Contrasting Layers Improves Factuality in Large Language Models},
    author={Yung-Sung Chuang and Yujia Xie and Hongyin Luo and Yoon Kim and James R. Glass and Pengcheng He},
    booktitle={The Twelfth International Conference on Learning Representations},
    year={2024},
    url={https://openreview.net/forum?id=Th6NyL07na}
}

@inproceedings{leng2024mitigating,
  title={Mitigating object hallucinations in large vision-language models through visual contrastive decoding},
  author={Leng, Sicong and Zhang, Hang and Chen, Guanzheng and Li, Xin and Lu, Shijian and Miao, Chunyan and Bing, Lidong},
  booktitle={Proceedings of the IEEE/CVF Conference on Computer Vision and Pattern Recognition},
  pages={13872--13882},
  year={2024}
}

@inproceedings{liu2025paying,
  title={Paying more attention to image: A training-free method for alleviating hallucination in lvlms},
  author={Liu, Shi and Zheng, Kecheng and Chen, Wei},
  booktitle={European Conference on Computer Vision},
  pages={125--140},
  year={2025},
  organization={Springer}
}

@article{woo2024don,
  title={Don't Miss the Forest for the Trees: Attentional Vision Calibration for Large Vision Language Models},
  author={Woo, Sangmin and Kim, Donguk and Jang, Jaehyuk and Choi, Yubin and Kim, Changick},
  journal={arXiv preprint arXiv:2405.17820},
  year={2024}
}

@inproceedings{favero2024multi,
  title={Multi-modal hallucination control by visual information grounding},
  author={Favero, Alessandro and Zancato, Luca and Trager, Matthew and Choudhary, Siddharth and Perera, Pramuditha and Achille, Alessandro and Swaminathan, Ashwin and Soatto, Stefano},
  booktitle={Proceedings of the IEEE/CVF Conference on Computer Vision and Pattern Recognition},
  pages={14303--14312},
  year={2024}
}

@article{yu2023evaluating,
  title={Evaluating progress in automatic chest x-ray radiology report generation},
  author={Yu, Feiyang and Endo, Mark and Krishnan, Rayan and Pan, Ian and Tsai, Andy and Reis, Eduardo Pontes and Fonseca, Eduardo Kaiser Ururahy Nunes and Lee, Henrique Min Ho and Abad, Zahra Shakeri Hossein and Ng, Andrew Y and others},
  journal={Patterns},
  volume={4},
  number={9},
  year={2023},
  publisher={Elsevier}
}

@inproceedings{smit2020combining,
  title={Combining Automatic Labelers and Expert Annotations for Accurate Radiology Report Labeling Using BERT},
  author={Smit, Akshay and Jain, Saahil and Rajpurkar, Pranav and Pareek, Anuj and Ng, Andrew Y and Lungren, Matthew},
  booktitle={Proceedings of the 2020 Conference on Empirical Methods in Natural Language Processing (EMNLP)},
  pages={1500--1519},
  year={2020}
}

@inproceedings{zhao2024ratescore,
  title={RaTEScore: A Metric for Radiology Report Generation},
  author={Zhao, Weike and Wu, Chaoyi and Zhang, Xiaoman and Zhang, Ya and Wang, Yanfeng and Xie, Weidi},
  booktitle={Proceedings of the 2024 Conference on Empirical Methods in Natural Language Processing},
  pages={15004--15019},
  year={2024}
}

@inproceedings{gong2024damro,
  title={DAMRO: Dive into the Attention Mechanism of LVLM to Reduce Object Hallucination},
  author={Gong, Xuan and Ming, Tianshi and Wang, Xinpeng and Wei, Zhihua},
  booktitle={Proceedings of the 2024 Conference on Empirical Methods in Natural Language Processing},
  pages={7696--7712},
  year={2024}
}

@inproceedings{yuan2024helpd,
  title={HELPD: Mitigating Hallucination of LVLMs by Hierarchical Feedback Learning with Vision-enhanced Penalty Decoding},
  author={Yuan, Fan and Qin, Chi and Xu, Xiaogang and Li, Piji},
  booktitle={Proceedings of the 2024 Conference on Empirical Methods in Natural Language Processing},
  pages={1768--1785},
  year={2024}
}

@inproceedings{liang2024mitigating,
  title={Mitigating Hallucination in Visual-Language Models via Re-balancing Contrastive Decoding},
  author={Liang, Xiaoyu and Yu, Jiayuan and Mu, Lianrui and Zhuang, Jiedong and Hu, Jiaqi and Yang, Yuchen and Ye, Jiangnan and Lu, Lu and Chen, Jian and Hu, Haoji},
  booktitle={Chinese Conference on Pattern Recognition and Computer Vision (PRCV)},
  pages={482--496},
  year={2024},
  organization={Springer}
}

@inproceedings{
darcet2024vision,
title={Vision Transformers Need Registers},
author={Timoth{\'e}e Darcet and Maxime Oquab and Julien Mairal and Piotr Bojanowski},
booktitle={The Twelfth International Conference on Learning Representations},
year={2024},
url={https://openreview.net/forum?id=2dnO3LLiJ1}
}

@article{zhang2023biomedclip,
  title={BiomedCLIP: a multimodal biomedical foundation model pretrained from fifteen million scientific image-text pairs},
  author={Zhang, Sheng and Xu, Yanbo and Usuyama, Naoto and Xu, Hanwen and Bagga, Jaspreet and Tinn, Robert and Preston, Sam and Rao, Rajesh and Wei, Mu and Valluri, Naveen and others},
  journal={arXiv preprint arXiv:2303.00915},
  year={2023}
}

@inproceedings{
hu2021lora,
title={Lo{RA}: Low-Rank Adaptation of Large Language Models},
author={Edward J Hu and yelong shen and Phillip Wallis and Zeyuan Allen-Zhu and Yuanzhi Li and Shean Wang and Lu Wang and Weizhu Chen},
booktitle={International Conference on Learning Representations},
year={2022},
url={https://openreview.net/forum?id=nZeVKeeFYf9}
}

@inproceedings{liu2021slake,
  title={Slake: A semantically-labeled knowledge-enhanced dataset for medical visual question answering},
  author={Liu, Bo and Zhan, Li-Ming and Xu, Li and Ma, Lin and Yang, Yan and Wu, Xiao-Ming},
  booktitle={2021 IEEE 18th International Symposium on Biomedical Imaging (ISBI)},
  pages={1650--1654},
  year={2021},
  organization={IEEE}
}

@article{lau2018dataset,
  title={A dataset of clinically generated visual questions and answers about radiology images},
  author={Lau, Jason J and Gayen, Soumya and Ben Abacha, Asma and Demner-Fushman, Dina},
  journal={Scientific data},
  volume={5},
  number={1},
  pages={1--10},
  year={2018},
  publisher={Nature Publishing Group}
}

@article{he2020pathvqa,
  title={Pathvqa: 30000+ questions for medical visual question answering},
  author={He, Xuehai and Zhang, Yichen and Mou, Luntian and Xing, Eric and Xie, Pengtao},
  journal={arXiv preprint arXiv:2003.10286},
  year={2020}
}

@article{demner2016preparing,
  title={Preparing a collection of radiology examinations for distribution and retrieval},
  author={Demner-Fushman, Dina and Kohli, Marc D and Rosenman, Marc B and Shooshan, Sonya E and Rodriguez, Laritza and Antani, Sameer and Thoma, George R and McDonald, Clement J},
  journal={Journal of the American Medical Informatics Association},
  volume={23},
  number={2},
  pages={304--310},
  year={2016},
  publisher={Oxford University Press}
}

@inproceedings{hu2024omnimedvqa,
  title={Omnimedvqa: A new large-scale comprehensive evaluation benchmark for medical lvlm},
  author={Hu, Yutao and Li, Tianbin and Lu, Quanfeng and Shao, Wenqi and He, Junjun and Qiao, Yu and Luo, Ping},
  booktitle={Proceedings of the IEEE/CVF Conference on Computer Vision and Pattern Recognition},
  pages={22170--22183},
  year={2024}
}

@article{johnson2019mimic,
  title={MIMIC-CXR-JPG, a large publicly available database of labeled chest radiographs},
  author={Johnson, Alistair EW and Pollard, Tom J and Greenbaum, Nathaniel R and Lungren, Matthew P and Deng, Chih-ying and Peng, Yifan and Lu, Zhiyong and Mark, Roger G and Berkowitz, Seth J and Horng, Steven},
  journal={arXiv preprint arXiv:1901.07042},
  year={2019}
}

@article{xia2024cares,
  title={Cares: A comprehensive benchmark of trustworthiness in medical vision language models},
  author={Xia, Peng and Chen, Ze and Tian, Juanxi and Gong, Yangrui and Hou, Ruibo and Xu, Yue and Wu, Zhenbang and Fan, Zhiyuan and Zhou, Yiyang and Zhu, Kangyu and others},
  journal={Advances in Neural Information Processing Systems},
  volume={37},
  pages={140334--140365},
  year={2024}
}

@inproceedings{banerjee2005meteor,
  title={METEOR: An automatic metric for MT evaluation with improved correlation with human judgments},
  author={Banerjee, Satanjeev and Lavie, Alon},
  booktitle={Proceedings of the acl workshop on intrinsic and extrinsic evaluation measures for machine translation and/or summarization},
  pages={65--72},
  year={2005}
}

@inproceedings{papineni2002bleu,
  title={Bleu: a method for automatic evaluation of machine translation},
  author={Papineni, Kishore and Roukos, Salim and Ward, Todd and Zhu, Wei-Jing},
  booktitle={Proceedings of the 40th annual meeting of the Association for Computational Linguistics},
  pages={311--318},
  year={2002}
}

@inproceedings{
zhang2019bertscore,
title={BERTScore: Evaluating Text Generation with BERT},
author={Tianyi Zhang and Varsha Kishore and Felix Wu and Kilian Q. Weinberger and Yoav Artzi},
booktitle={International Conference on Learning Representations},
year={2020},
url={https://openreview.net/forum?id=SkeHuCVFDr}
}

@inproceedings{lin2004rouge,
  title={Rouge: A package for automatic evaluation of summaries},
  author={Lin, Chin-Yew},
  booktitle={Text summarization branches out},
  pages={74--81},
  year={2004}
}

@inproceedings{
chexagent-2024,
title={CheXagent: Towards a Foundation Model for Chest X-Ray Interpretation},
author={Zhihong Chen and Maya Varma and Jean-Benoit Delbrouck and Magdalini Paschali and Louis Blankemeier and Dave Van Veen and Jeya Maria Jose Valanarasu and Alaa Youssef and Joseph Paul Cohen and Eduardo Pontes Reis and Emily Tsai and Andrew Johnston and Cameron Olsen and Tanishq Mathew Abraham and Sergios Gatidis and Akshay S Chaudhari and Curtis Langlotz},
booktitle={AAAI 2024 Spring Symposium on Clinical Foundation Models},
year={2024},
url={https://openreview.net/forum?id=P3LOmrZWGR}
}

@article{liu2024visual,
  title={Visual instruction tuning},
  author={Liu, Haotian and Li, Chunyuan and Wu, Qingyang and Lee, Yong Jae},
  journal={Advances in neural information processing systems},
  volume={36},
  year={2024}
}

@inproceedings{
Holtzman2020The,
title={The Curious Case of Neural Text Degeneration},
author={Ari Holtzman and Jan Buys and Li Du and Maxwell Forbes and Yejin Choi},
booktitle={International Conference on Learning Representations},
year={2020},
url={https://openreview.net/forum?id=rygGQyrFvH}
}

@inproceedings{NIPS2014_beam,
 author = {Sutskever, Ilya and Vinyals, Oriol and Le, Quoc V},
 booktitle = {Advances in Neural Information Processing Systems},
 editor = {Z. Ghahramani and M. Welling and C. Cortes and N. Lawrence and K.Q. Weinberger},
 pages = {},
 publisher = {Curran Associates, Inc.},
 title = {Sequence to Sequence Learning with Neural Networks},
 url = {https://proceedings.neurips.cc/paper_files/paper/2014/file/a14ac55a4f27472c5d894ec1c3c743d2-Paper.pdf},
 volume = {27},
 year = {2014}
}

@article{gu2024medvh,
  title={MedVH: Towards Systematic Evaluation of Hallucination for Large Vision Language Models in the Medical Context},
  author={Gu, Zishan and Yin, Changchang and Liu, Fenglin and Zhang, Ping},
  journal={arXiv preprint arXiv:2407.02730},
  year={2024}
}

@article{chen2024detecting,
  title={Detecting and Evaluating Medical Hallucinations in Large Vision Language Models},
  author={Chen, Jiawei and Yang, Dingkang and Wu, Tong and Jiang, Yue and Hou, Xiaolu and Li, Mingcheng and Wang, Shunli and Xiao, Dongling and Li, Ke and Zhang, Lihua},
  journal={arXiv preprint arXiv:2406.10185},
  year={2024}
}

@inproceedings{Jiang2024CoMTCR,
  title={CoMT: Chain-of-Medical-Thought Reduces Hallucination in Medical Report Generation},
  author={Yue Jiang and Jiawei Chen and Dingkang Yang and Mingcheng Li and Shunli Wang and Tong Wu and Ke Li and Lihua Zhang},
  year={2024},
  url={https://api.semanticscholar.org/CorpusID:270559982}
}

@article{jain2021radgraph,
  title={Radgraph: Extracting clinical entities and relations from radiology reports},
  author={Jain, Saahil and Agrawal, Ashwin and Saporta, Adriel and Truong, Steven QH and Duong, Du Nguyen and Bui, Tan and Chambon, Pierre and Zhang, Yuhao and Lungren, Matthew P and Ng, Andrew Y and others},
  journal={arXiv preprint arXiv:2106.14463},
  year={2021}
}

@article{khattak2024unimed,
  title={UniMed-CLIP: Towards a Unified Image-Text Pretraining Paradigm for Diverse Medical Imaging Modalities},
  author={Khattak, Muhammad Uzair and Kunhimon, Shahina and Naseer, Muzammal and Khan, Salman and Khan, Fahad Shahbaz},
  journal={arXiv preprint arXiv:2412.10372},
  year={2024}
}

@inproceedings{han2023word,
  title={Word Embeddings Are Steers for Language Models},
  author={Han, Chi and Xu, Jialiang and Li, Manling and Fung, Yi and Sun, Chenkai and Jiang, Nan and Abdelzaher, Tarek and Ji, Heng},
  booktitle={Proceedings of the 62nd Annual Meeting of the Association for Computational Linguistics (Volume 1: Long Papers)},
  pages={16410--16430},
  year={2024}
}

@inproceedings{jiang2024devils,
  title={Devils in middle layers of large vision-language models: Interpreting, detecting and mitigating object hallucinations via attention lens},
  author={Jiang, Zhangqi and Chen, Junkai and Zhu, Beier and Luo, Tingjin and Shen, Yankun and Yang, Xu},
  booktitle={Proceedings of the Computer Vision and Pattern Recognition Conference},
  pages={25004--25014},
  year={2025}
}

@article{neo2024towards,
  title={Towards interpreting visual information processing in vision-language models},
  author={Neo, Clement and Ong, Luke and Torr, Philip and Geva, Mor and Krueger, David and Barez, Fazl},
  journal={arXiv preprint arXiv:2410.07149},
  year={2024}
}

@article{tu2025attention,
  title={Attention Reallocation: Towards Zero-cost and Controllable Hallucination Mitigation of MLLMs},
  author={Tu, Chongjun and Ye, Peng and Zhou, Dongzhan and Bai, Lei and Yu, Gang and Chen, Tao and Ouyang, Wanli},
  journal={arXiv preprint arXiv:2503.08342},
  year={2025}
}

@article{chen2025attention,
  title={Attention Hijackers: Detect and Disentangle Attention Hijacking in LVLMs for Hallucination Mitigation},
  author={Chen, Beitao and Lyu, Xinyu and Gao, Lianli and Song, Jingkuan and Shen, Heng Tao},
  journal={arXiv preprint arXiv:2503.08216},
  year={2025}
}

@inproceedings{chang2025focus,
    title = "Focus on What Matters: Enhancing Medical Vision-Language Models with Automatic Attention Alignment Tuning",
    author = "Chang, Aofei  and
      Huang, Le  and
      Boyd, Alex James  and
      Bhatia, Parminder  and
      Kass-Hout, Taha  and
      Xiao, Cao  and
      Ma, Fenglong",
    booktitle = "Proceedings of the 63rd Annual Meeting of the Association for Computational Linguistics (Volume 1: Long Papers)",
    month = jul,
    year = "2025",
    address = "Vienna, Austria",
    publisher = "Association for Computational Linguistics",
    url = "https://aclanthology.org/2025.acl-long.460/",
    pages = "9357--9372",
}

@inproceedings{wang2024recent,
  title={Recent advances in predictive modeling with electronic health records},
  author={Wang, Jiaqi and Luo, Junyu and Ye, Muchao and Wang, Xiaochen and Zhong, Yuan and Chang, Aofei and Huang, Guanjie and Yin, Ziyi and Xiao, Cao and Sun, Jimeng and others},
  booktitle={IJCAI: proceedings of the conference},
  volume={2024},
  pages={8272},
  year={2024}
}

\appendix

\newpage


\section{Appendix}

\section{A. Limitations}
\label{limitations}
While \ours improves visual representation and attention alignment in Med-LVLMs, it has several limitations:
(1) The method assumes access to a high-quality expert CLIP model, and its effectiveness depends on the domain relevance and quality of this encoder.
(2) Our approach performs alignment through lightweight distillation at a single layer. While this design ensures efficiency, stronger performance may be achieved with end-to-end optimization using domain-specific CLIP models and larger datasets—albeit at the cost of increased complexity and computational resources.
(3) Although we address token mismatches via interpolation, aligning features across different resolutions or architectures may still introduce noise or minor inconsistencies.

\section{B. Dataset Processing and Statistics}
\label{appd:dataset}

For IU-Xray and OmniVQA in the medical VQA task, we use the preprocessed datasets provided by the CARES benchmark~\cite{xia2024cares}, splitting each dataset into training and test sets with a 7:3 ratio based on image cases to avoid label leakage. For the IU-Xray dataset used in report generation, we perform no additional processing and directly utilize the publicly available version. For the MIMIC-CXR dataset used in the report generation task, we randomly sample 2,000 image-report pairs from the preprocessed MIMIC-CXR-JPG dataset~\cite{johnson2019mimic} for the training set and 500 pairs for the test set. We extract the ``Findings'' and ``Impression'' sections from each report in sampled MIMIC-CXR reports, filtering out those with an extremely low word count. The statistics of those datasets are shown in Table~\ref{tab:dataset_stats}. Unless otherwise specified, we use 10\% of the training set for validation.

\begin{table}[h]
\centering
\resizebox{0.8\columnwidth}{!}{
\begin{tabular}{c|c|c|c}
\hline
\textbf{Task}          & \textbf{Dataset} & \textbf{Training} & \textbf{Test} \\ \hline
\multirow{5}{*}{\makecell{Medical \\ VQA} } 
& SLAKE           & 4,919                     & 1,061        \\ \cline{2-4} 
& VQA-RAD         & 1,797                     & 451  \\ \cline{2-4} 
& PathVQA         & 19,755                     & 6,761    \\ 
\cline{2-4} 
& IU-Xray         & 1,789                     & 784     \\ 
\cline{2-4} 
& OmniMedVQA         & 6,155                     & 2,642  \\ 
\hline
\multirow{2}{*}{\makecell{Report \\ Generation} } 
& IU-Xray         & 2,069                     & 590        \\ \cline{2-4} 
& MIMIC-CXR       & 1,902                     & 441       \\ \hline
\end{tabular}}
\caption{Dataset statistics for the Medical VQA and Report Generation tasks.}
\label{tab:dataset_stats}
\vspace{-0.2in}
\end{table}

\section{C. Implementation Details of Baselines}
\label{appd:implementation_baselines}
Generally, we follow the recommended settings for all baselines while making necessary adjustments to adapt them to Med-LVLMs. The detailed settings are listed as follows:
\begin{itemize}  
    \item Beam Search~\cite{NIPS2014_beam}: The number of beams is set to 5.  
    \item Nucleus Sampling~\cite{Holtzman2020The}: The top-\( p \) value for sampling is \( 0.9 \).  
    \item VCD~\cite{leng2024mitigating}: The contrastive decoding parameters are set to \( \alpha = 1 \) and \( \beta = 0.1 \). Diffusion noise is added to images using 500 steps.  
    \item DoLa~\cite{chuang2023dola}: The mature layer is set to 32, while the early candidate mature layers are \([0,2,4,6,8,10,12,14]\).  
    \item OPERA~\cite{huang2024opera}: The number of beams is set to 5, with a scale factor of 50, threshold of 15, and \(\text{num-attn-candidates} = 5\). The penalty weight is set to 1. Notably, for LLaVA-Med-1.5 in the report generation task, the scale factor is set to 25 and the threshold is adjusted to 25, as the default values result in nonsensical decoded content.  
    \item AVISC~\cite{woo2024don}: We select the top-10 outlier image tokens to construct the negative decoding object. The contrastive decoding parameters are set to \( \alpha = 1 \) and \( \beta = 0.1 \).
    \item M3ID~\cite{favero2024multi}: The contrastive decoding parameters are set as follows: $\lambda = 0.02$ and $\gamma_t = \exp(-\lambda \cdot t)$, where $t$ denotes the current decoding step.
    \item DAMRO~\cite{gong2024damro}: We select the top-10 tokens with the highest attention to the [CLS] token in the visual encoder as outlier tokens. The contrastive decoding parameters are set to \( \alpha = 0.5 \) and \( \beta = 0.1 \).
    \item PAI~\cite{liu2025paying}: In the inference intervention, the start layer and end layer are set to 2 and 32, respectively, \(\gamma = 1.1\) and \(\alpha = 0.2\).
\end{itemize}  

\section{D. Implementation Details}
\label{appd:implementation}

\subsection{D.1 Hyperparameter setting}
All fine-tuning experiments are conducted with a fixed random seed for LoRA initialization to ensure reproducibility. We use a LoRA rank of 64 for the report generation task and 32 for medical VQA tasks. All experiments are run on a system with four NVIDIA A6000 GPUs, using Ubuntu 22.04.1 and CUDA version 12.6. We apply our distillation losses at layer 20 in LLaVA-Med-1.5 (total 32 layers) and layer 16 in HuatuoGPT-Vision-7B (total 28 layers).

\subsection{D.2 Training Details and $\alpha$ and $\beta$ Selection}
In the final loss formulation (Eq.\eqref{eq:final}), two hyperparameters, $\alpha$ and $\beta$, are introduced to balance the alignment objectives with the original LLM loss. A detailed analysis of these hyperparameters is provided in Appendix~\ref{appd:hyperparameter}. In our experiments, we consistently set $\alpha = 1$ on almost all datasets, which yields strong performance across tasks; exploring more advanced tuning strategies for $\alpha$ is left for future work. We tune $\beta$ to control the degree of attention modulation. The chosen values of $\beta$ and the corresponding fine-tuning epochs for each dataset are listed in Table~\ref{tab:hyperparams}.


\begin{table}[h]
\centering
\vspace{-0.1in}
    \centering
    \resizebox{1\columnwidth}{!}{
    \begin{tabular}{l|c|cc|cc}
        \toprule
        \multirow{2}{*}{\textbf{Dataset}} & \multirow{2}{*}{\textbf{Epochs}} 
        & \multicolumn{2}{c|}{LLaVA-Med-1.5} 
        & \multicolumn{2}{c}{HuatuoGPT-Vision} \\
        \cline{3-6}
        & & \textbf{$\alpha$} & \textbf{$\beta$} & \textbf{$\alpha$} & \textbf{$\beta$} \\
        \midrule
        
        \multicolumn{6}{c}{\textit{Medical Report Generation}} \\
        \midrule
        IU-Xray       & 12 & 1.0 & 0.03 & 1.0 & 0.03 \\
        MIMIC-CXR     & 12 & 1.0 & 0.03 & 1.0 & 0.03 \\
        \midrule
        \multicolumn{6}{c}{\textit{Medical VQA}} \\
        \midrule
        SLAKE         & 6  & 1.0 & 0.03 & 1.0 & 0.1 \\
        VQA-RAD       & 9  & 1.0 & 0.06 & 1.0 & 0.1 \\
        PathVQA       & 3  & 0.1 & 0.01 & 0.1 & 0.01 \\
        IU-Xray       & 6  & 1.0 & 0.1 & 1.0 & 0.2 \\
        OmniMedVQA    & 3  & 1.0 & 0.02 & 1.0 & 0.02 \\
        
        \bottomrule
    \end{tabular}}
    \vspace{-0.1in}
    \caption{Fine-tuning epochs and loss weights for different datasets.}
    \label{tab:hyperparams}
\end{table}

\subsection{D.3 Details and Examples of Interpolation}
\label{appd:interpolation}
To enable alignment distillation from expert CLIP models that produce a different number of visual tokens compared to the target Med-LVLM, we apply interpolation to both visual features and attention maps. This step ensures compatibility in spatial resolution and allows token-wise alignment in our proposed framework.

Different CLIP models tokenize images at varying resolutions and patch sizes. For instance, UniMed-CLIP (ViT-B/16, 224px) produces $M = 14 \times 14 = 196$ tokens, while LLaVA-Med-1.5 and HuatuoGPT-Vision-7B typically use $N = 24 \times 24 = 576$ tokens. To distill alignment knowledge (e.g., similarity structure or attention distribution), we must first transform the expert outputs to match the Med-LVLM’s token resolution.

\paragraph{Interpolation on Visual Features.}
Given expert visual features $\mathbf{E}_v \in \mathbb{R}^{M \times b}$, we reshape them into a 2D spatial grid $\mathbb{R}^{H \times W \times b}$ with $M = H \times W$, apply bilinear interpolation to resize to the target grid size $\mathbb{R}^{H' \times W' \times b}$ where $N = H' \times W'$, and flatten to obtain interpolated features $\tilde{\mathbf{E}}_v \in \mathbb{R}^{N \times b}$. These features are then used to compute the expert similarity matrix $\mathbf{S}^e \in \mathbb{R}^{N \times N}$.

\paragraph{Interpolation on Attention Maps.}
Similarly, for attention maps $\mathbf{E}_a \in \mathbb{R}^{M}$ from expert CLIP models, we reshape them into a $H \times W$ grid, interpolate to $H' \times W'$ using bilinear interpolation, and flatten to obtain the target attention map $\tilde{\mathbf{E}}_a \in \mathbb{R}^N$.

\begin{figure}
    \centering
    \includegraphics[width=1\columnwidth]{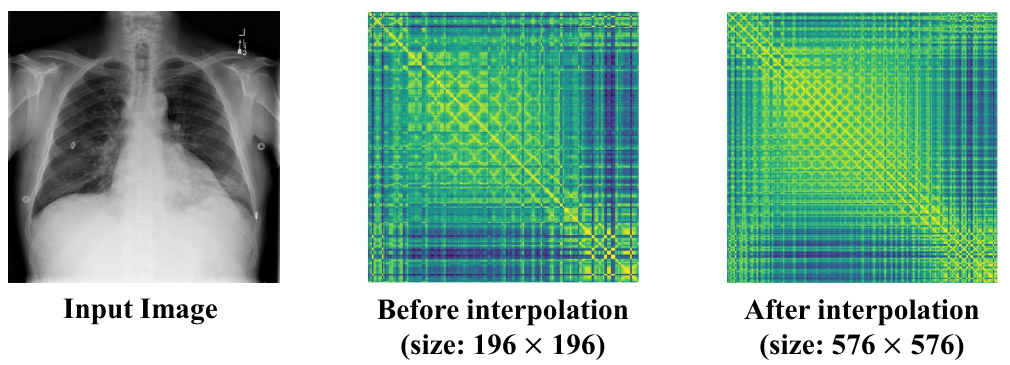}
    \caption{Example of similarity matrices before and after interpolation of visual features.}
    \label{fig:appd_interp_sim}
    \vspace{-0.2in}
\end{figure}

\begin{figure}
    \centering
    \vspace{-12pt}
    \includegraphics[width=1\columnwidth]{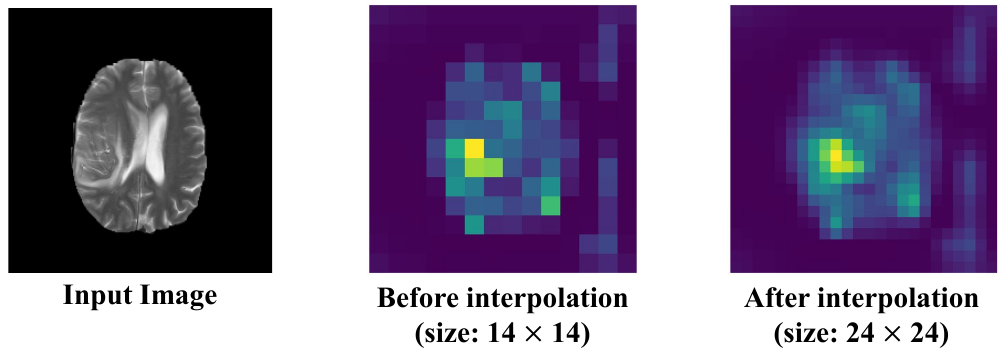}
    \caption{Example of attention maps before and after interpolation.}
    \label{fig:appd_interp_attn}
    \vspace{-0.2in}
\end{figure}

\paragraph{Interpolation Implementation.}
We use PyTorch's standard bilinear interpolation function \texttt{torch.nn.functional.interpolate} to resize 2D spatial grids. For feature maps, we reshape the expert visual features to shape $[1, b, H, W]$ (channels-first format), apply bilinear interpolation to the target size $[H', W']$, and then flatten back to $[N, b]$ where $N = H' \times W'$. For attention maps, which are single-channel vectors, we reshape to $[1, 1, H, W]$, interpolate, and flatten similarly. This approach ensures consistency between visual feature resizing and attention map resizing. All interpolated outputs are renormalized if necessary (e.g., via $\ell_2$ normalization for feature vectors).

\paragraph{Examples.}
We provide qualitative examples of similarity matrices and attention maps before and after interpolation in Figure~\ref{fig:appd_interp_sim} and Figure~\ref{fig:appd_interp_attn}. The expert model used here is BiomedCLIP~\cite{zhang2023biomedclip}, which generates $196 = 14 \times 14$ visual tokens. As shown, our interpolation approach effectively preserves both spatial and semantic structures, enabling meaningful alignment supervision even under differing token resolutions.

\section{E. Metrics}
\label{appd:metrics}
\subsection{E.1 Metrics for Report Generation}
\label{appd:metric_report}
We evaluate model performance using commonly used metrics for generation tasks. These include BERTScore~\cite{zhang2019bertscore}, which measures the similarity between the embeddings of predicted and reference texts, and METEOR~\cite{banerjee2005meteor}, which evaluates alignment between generated answers and reference texts, accounting for synonyms and stemming. Additionally, we employ ROUGE-L~\citep{lin2004rouge}, which measures n-gram overlap and the longest common subsequence, and BLEU~\cite{papineni2002bleu}, which calculates n-gram precision in the predicted text relative to the reference, focusing on exact matches. In addition, we include the following domain-specific metrics designed for medical report generation: 
\begin{itemize}
    \item CheXbert~\cite{smit2020combining} is an automatic labeler that extracts pathology indicators from radiology reports. We follow~\cite{yu2023evaluating} to calculate the CheXbert vector similarity that measures the cosine similarity between pathology indicator vectors derived from ground truth and model-generated reports. 
    \item RadGraph~\cite{jain2021radgraph} is a tool that extracts entity and relation from radiology reports. We use RadGraph to specifically indicate RadGraph F1, which measures the overlap of clinical entities and their relations extracted from ground truth and model-generated reports.
    \item RaTEScore~\cite{zhao2024ratescore} is a recently proposed metric that prioritizes crucial medical entities, including diagnostic outcomes and anatomical details. This metric is robust to complex medical synonyms and sensitive to negation expressions, aligning more closely with human judgment compared to existing metrics.
\end{itemize}

\begin{table*}[t]
\centering

\resizebox{1.8\columnwidth}{!}{
\begin{tabular}{c|c|ccccccc}
\toprule
\multirow{2}{*}{\textbf{Model}} & \multirow{2}{*}{\textbf{Method}} & \multicolumn{7}{c}{\textbf{Metrics}}\\
\cline{3-9}
 & & \textbf{BLEU} & \textbf{\text{ROUGE-L}} & \textbf{METEOR} & \textbf{BERTScore} & \textbf{CheXbert} & \textbf{RadGraph} & \textbf{RaTEScore}\\
\midrule
\multirow{11}{*}{\textbf{IU-Xray}}
& Greedy  & 1.04 & 12.15 & 9.87 & 85.43 & 38.04 & 5.43 & 34.97   \\
& Beam & 1.09 & 11.17 & 19.59 & 83.43 & 40.13 & 9.42 & 48.64 \\
& Nucleus & 1.44 & 12.10 & 15.60 & 81.45 & 38.04 & 6.51 & 40.41 \\
& VCD & 1.42 & 12.29 & 15.72 & 84.54 & 36.57 & 6.49 & 39.93 \\
& DoLa & 0.99 & 12.15 & 9.36 & 85.64 & 38.22 & 5.40 & 34.93  \\
& OPERA & 1.13 & 11.49 & 14.63 & 83.76 & 37.38 & 1.41 & 35.96 \\
& AVISC & 1.18 & 11.32 & 16.66 & 83.76 & 35.83 & 6.63 & 40.36  \\
& M3ID & 1.33 & 12.31 & 16.31 & 84.45 & 37.54 & 6.42 & 40.41 \\
& DAMRO & 1.27 & 11.56 & 16.42 & 84.08 & 35.60 & 6.80 & 40.08 \\
& PAI & 1.11 & 12.05 & 10.99 & 85.03 & 37.56 & 5.21 & 34.83 \\
& A$^3$Tune & 10.51 & 28.76 & 35.74 & 88.51 & 53.88 & 23.10 & 59.66 \\
& \cellcolor{myhighlight} \ours & \cellcolor{myhighlight}10.31 & \cellcolor{myhighlight}29.01 & \cellcolor{myhighlight}35.22 & \cellcolor{myhighlight}88.66 & \cellcolor{myhighlight}55.62 & \cellcolor{myhighlight}23.29 & \cellcolor{myhighlight}59.99 \\

\midrule

\multirow{11}{*}{\textbf{MIMIC-CXR}}
& Greedy  & 0.93 & 10.90 & 9.45 & 83.09 & 14.11 & 1.09 & 28.07   \\
& Beam & 1.13 & 10.95 & 13.01 & 82.34 & 12.51 & 1.87 & 32.86 \\
& Nucleus & 0.96 & 10.55 & 10.70 & 78.53 & 14.11 & 1.64 & 31.54 \\
& VCD & 1.00 & 10.93 & 11.31 & 83.11 & 13.46 & 1.97 & 31.75 \\
& DoLa & 0.65 & 9.88 & 7.83 & 83.47 & 14.07 & 1.09 & 28.07  \\
& OPERA & 1.09 & 11.51 & 12.22 & 82.69 & 13.00 & 0.52 & 27.77 \\
& AVISC & 1.16 & 11.14 & 12.30 & 82.61 & 14.11 & 1.97 & 32.20  \\
& M3ID & 1.06 & 11.50 & 11.54 & 83.15 & 13.98 & 2.24 & 32.74 \\
& DAMRO & 1.14 & 11.01 & 12.35 & 82.80 & 13.73 & 2.26 & 32.18 \\
& PAI & 1.12 & 11.67 & 10.63 & 82.80 & 14.05 & 0.98 & 28.07 \\
& A$^3$Tune & 4.22 & 18.02 & 20.69 & 85.75 & 25.37 & 10.52 & 42.15 \\
& \cellcolor{myhighlight} \ours & \cellcolor{myhighlight}4.51 & \cellcolor{myhighlight}18.43 & \cellcolor{myhighlight}20.80 & \cellcolor{myhighlight}86.00 & \cellcolor{myhighlight}25.67 & \cellcolor{myhighlight}10.92 & \cellcolor{myhighlight}42.03 \\

\bottomrule
\end{tabular}}

\caption{Baseline Results on report generation benchmarks without LoRA-Fine-tuning, based on LLaVA-Med-1.5. }
\label{tab:report_original_llavamed1.5}
\vspace{-0.1in}
\end{table*}

\begin{figure}
    \centering
    \includegraphics[width=1\columnwidth]{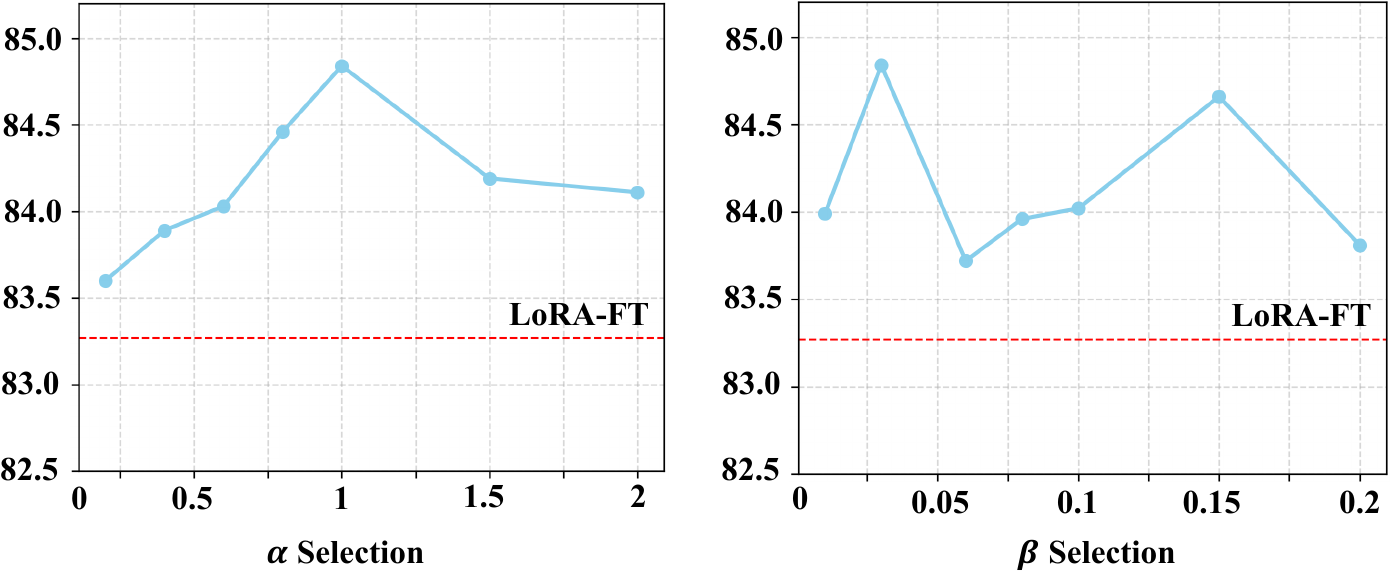}
    \caption{Results of hyperparameter analysis. Red lines denote the baseline results of LoRA tuning.}
    \label{fig:hyper_appendix}
\end{figure}


\begin{table*}[h!]
\centering
\resizebox{1.8\columnwidth}{!}{
\begin{tabular}{c|c|ccccccc}
\toprule
\multirow{2}{*}{\textbf{Model}} & \multirow{2}{*}{\textbf{Expert CLIP}} & \multicolumn{7}{c}{\textbf{Metrics}}\\
\cline{3-9}
 & & \textbf{BLEU} & \textbf{\text{ROUGE-L}} & \textbf{METEOR} & \textbf{BERTScore} & \textbf{CheXbert} & \textbf{RadGraph} & \textbf{RaTEScore}\\
\midrule
\multirow{3}{*}{\textbf{\makecell{HuatuoGPT-Vision\\ + LoRA}}}
& Biomed  & 11.01 & 29.97 & 33.72 & 88.89 & 56.06 & 24.09 & 60.63   \\
& UniMed-B & 10.12 & 28.80 & 34.54 & 88.75 & 55.25 & 23.50 & 60.61 \\
& UniMed-L & 10.73 & 29.10 & 36.30 & 88.67 & 56.27 & 23.51 & 60.49 \\

\midrule

\multirow{3}{*}{\textbf{\makecell{LLaVA-Med-1.5 \\ + LoRA}}}
& Biomed  & 10.26 & 28.56 & 35.15 & 88.54 & 54.93 & 22.93 & 59.20   \\
& UniMed-B & 10.05 & 28.97 & 34.73 & 88.76 & 55.87 & 23.24 & 61.00 \\
& UniMed-L & 10.31 & 29.01 & 35.22 & 88.66 & 55.62 & 23.29 & 59.99 \\

\bottomrule
\end{tabular}}
\vspace{-0.05in}
\caption{Performance of \ours using different expert CLIP models on the \textbf{IU-Xray} report generation benchmark, based on HuatuoGPT-Vision-7B and LLaVA-Med-1.5 with LoRA fine-tuning.}
\label{tab:more_clips_iu}
\vspace{-0.05in}
\end{table*}

\begin{table*}[h!]
\centering

\resizebox{1.8\columnwidth}{!}{
\begin{tabular}{c|c|ccccccc}
\toprule
\multirow{2}{*}{\textbf{Model}} & \multirow{2}{*}{\textbf{Expert CLIP}} & \multicolumn{7}{c}{\textbf{Metrics}}\\
\cline{3-9}
 & & \textbf{BLEU} & \textbf{\text{ROUGE-L}} & \textbf{METEOR} & \textbf{BERTScore} & \textbf{CheXbert} & \textbf{RadGraph} & \textbf{RaTEScore}\\
\midrule
\multirow{3}{*}{\textbf{\makecell{HuatuoGPT-Vision\\ + LoRA}}}
& Biomed  & 5.09 & 19.75 & 21.05 & 86.15 & 28.00 & 11.91 & 43.13   \\
& UniMed-B &  4.94 & 19.38 & 21.24 & 86.24 & 29.10 & 12.32 & 44.17 \\
& UniMed-L & 4.76 & 19.32 & 22.02 & 86.02 & 29.53 & 12.82 & 44.96 \\

\midrule

\multirow{3}{*}{\textbf{\makecell{LLaVA-Med-1.5 \\ + LoRA}}}
& Biomed  & 4.38 & 18.25 & 19.94 & 85.95 & 24.81 & 11.35 & 41.55   \\
& UniMed-B & 4.20 & 18.00 & 20.62 & 85.82 & 24.32 & 10.48 & 41.53 \\
& UniMed-L & 4.51 & 18.43 & 20.80 & 86.00 & 25.67 & 10.92 & 42.03 \\

\bottomrule
\end{tabular}}
\vspace{-0.05in}
\caption{Performance of \ours using different expert CLIP models on the \textbf{MIMIC-CXR} report generation benchmark, based on HuatuoGPT-Vision-7B and LLaVA-Med-1.5 with LoRA fine-tuning.}
\label{tab:more_clips_mimic}
\vspace{-0.05in}
\end{table*}

\begin{figure*}[t!]
  \centering
  \includegraphics[width=0.75\linewidth]{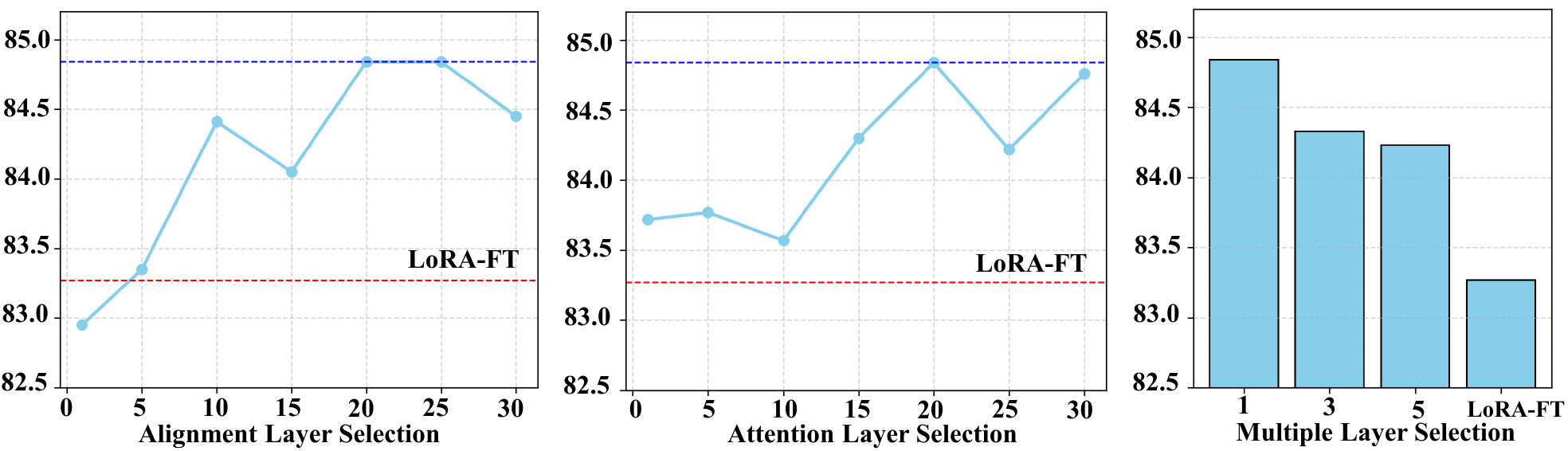}
  \caption{Results of layer selection. Blue lines denote the results where we use visual representation alignment and attention alignment on the same layer, layer 20.}
  \vspace{-0.1in}
  \label{fig:layer_selection_appendix}
\end{figure*}

\section{F. Additional Experiment Results}

\begin{table}[t]
\centering

\begin{tabular}{lccc}
\toprule
\textbf{Setting} & \makecell{\textbf{VQA-RAD} \\ \textbf{(open)}} & \makecell{\textbf{VQA-RAD} \\ \textbf{(close)}} & \makecell{\textbf{OmniMed} \\ \textbf{VQA}} \\
\midrule
\ours                 & \textbf{39.62} & \textbf{74.80} & \textbf{93.88} \\
\textsc{MedAlign}$_{\text{vis}}$ & 38.25 & 74.41 & 91.77 \\
\textsc{MedAlign}$_{\text{att}}$ & 37.36 & 74.80 & 91.66 \\
\midrule
LoRA-FT                   & 36.95 & 74.41 & 90.99 \\
w/o Distill               & 30.15 & 70.87 & 88.23 \\
\bottomrule
\end{tabular}
\caption{Ablation study across multiple datasets.}
\label{appd:ablation}
\vspace{-0.2in}
\end{table}

\subsection{F.1 Hyperparameter Analysis}
\label{appd:hyperparameter}
We conduct this analysis on the open-ended performance of the SLAKE dataset, with results shown in Figure~\ref{fig:hyper_appendix}.
 For the $\alpha$ selection, we fix $\beta = 0.03$ and vary the value of $\alpha$. The performance steadily improves as $\alpha$ increases, peaking at $\alpha = 1.0$, after which it begins to decline if $\alpha$ is increased further.
 For the $\beta$ selection, we fix $\alpha = 1.0$ and vary $\beta$. We observe that performance peaks at a small value of $\beta = 0.03$. Increasing $\beta$ beyond this point can still maintain relatively strong performance, as long as it does not become too large (e.g., $\beta = 0.2$ leads to a notable drop).

\subsection{F.2 Full Baseline Results on Report Generation}
\label{appd:full_baseline_results}

As discussed in the report generation results analysis, the original performance of baseline methods on Med-LVLMs is significantly low, as shown in Table~\ref{tab:report_original_llavamed1.5}. Therefore, we report only the fine-tuned baseline results in the main experiments for clearer and fair comparison. Additionally, we compare with the recent concurrent baseline A$^3$Tune~\cite{chang2025focus}, published in July, using the shared backbone LLaVA-Med-1.5. Although AAAI submission guidelines do not require comparisons with concurrent work, we include this for completeness. Notably, A$^3$Tune conducts its main experiments using the older LLaVA-Med backbone and reports results on LLaVA-Med-1.5 only in their appendix. For a fair and consistent comparison, we evaluate both methods solely on the shared LLaVA-Med-1.5 backbone. As shown in Table~\ref{tab:report_original_llavamed1.5}, \ours achieves further improvements across most report generation metrics. The performance gap can be attributed to the fundamental differences in approach. A$^3$Tune focuses on tuning attention distributions using weak segmentation labels, without addressing the underlying cause of suboptimal attention—namely, the misalignment between visual and textual modalities. In contrast, our method targets this deeper multimodal alignment issue and provides a more effective and interpretable solution.














\subsection{F.3 Additional Ablation Results}
\label{appd:ablation}
We conduct additional ablation studies on VQA-RAD and OmniMedVQA, as shown in Table~\ref{appd:ablation}. Removing either alignment loss leads to a performance drop, though the results remain above the LoRA-FT baseline. Furthermore, directly using UniMed-CLIP features without distillation results in degraded performance, likely due to feature mismatch. These findings further validate the effectiveness of our distillation design.

\subsection{F.4 Additional Results Using UniMed-CLIP ViT-B/16 (224px)}
\label{appd:full_results_unimed_base}

We report extended results on medical report generation using UniMed-CLIP ViT-B/16 (224px), denoted as UniMed-B, in Table~\ref{tab:more_clips_iu} and Table~\ref{tab:more_clips_mimic}, evaluated on both Med-LVLMs. Despite its lower resolution and smaller capacity compared to UniMed-L, UniMed-B still achieves strong performance. Notably, on LLaVA-Med-1.5 in Table~\ref{tab:more_clips_iu}, it obtains a RaTEScore of 61.00, even surpassing that of UniMed-L, demonstrating that even compact expert models can provide effective alignment guidance.

\subsection{F.5 Additional Results Using BiomedCLIP ViT-B/16 (224px)}
\label{appd:full_results_biomed}

We also present results using BiomedCLIP ViT-B/16 (224px), denoted as Biomed, in Table~\ref{tab:more_clips_iu} and Table~\ref{tab:more_clips_mimic}. While Biomed generally underperforms compared to UniMed-L, it still delivers competitive results. For instance, on HuatuoGPT-Vision in Table~\ref{tab:more_clips_iu}, Biomed achieves a RaTEScore of 60.63 and a BLEU score of 11.01, even higher than those obtained with UniMed-L, showcasing its potential as an alternative alignment source.

\subsection{F.6 Layer Selection Analysis}
\label{appd:layer_selection}


We present additional results on layer selection in Figure~\ref{fig:layer_selection_appendix}. For the alignment layer selection (visual feature alignment), where the attention distillation layer is fixed at layer 20, we observe that applying alignment at deeper layers improves performance up to a point and peaks before layer 25. Applying alignment too late (e.g., beyond layer 25) leads to degraded performance due to insufficient propagation of alignment signals. Similarly, for the attention layer selection, where the alignment layer is fixed at layer 20, we find that applying attention distillation at deeper layers also yields better results.

We also evaluate multi-layer distillation, testing configurations such as layers 19–21 (three-layer) and 18-22 (five-layer) ranges. Although performance remains strong, it slightly declines compared to using a single distillation layer. This suggests that aligning at one well-chosen intermediate layer is sufficient. As shown in our previous qualitative analysis, distillation at a single layer can propagate and enhance neighboring layers, validating the effectiveness and efficiency of our method.


\end{document}